\documentclass[10pt,journal,compsoc]{IEEEtran}
\ifCLASSOPTIONcompsoc
  \usepackage[nocompress]{cite}
\else
  \usepackage{cite}
\fi
\def\OURS{OOD-CV-v2~}

\usepackage[pagebackref,breaklinks,colorlinks]{hyperref}
\usepackage{comment}
\usepackage{pifont}
\usepackage{graphicx}
\usepackage{float}
\usepackage{tikz}
\usepackage{comment}
\usepackage{amsmath,amssymb} 
\usepackage{color}
\usepackage{booktabs}
\usepackage{bbding}
\usepackage{subcaption}
\usepackage{makecell}
\usepackage{color}
\usepackage{multirow}
\usepackage[capitalize]{cleveref}
\usepackage{soul}
\usepackage[final]{pdfpages}

\newcommand{\new}[1]{\textcolor{black}{#1}}
\newcommand{\zbc}[1]{\textcolor{black}{#1}}

\def\eg{\emph{e.g.}}

\definecolor{deepgreen}{rgb}{0,0.5,0}

\begin{document}

\title{\OURS: An extended Benchmark for Robustness to Out-of-Distribution Shifts of Individual Nuisances in Natural Images}

\author{Bingchen Zhao, Jiahao Wang, Wufei Ma, Artur Jesslen, Siwei Yang, Shaozuo Yu, \\Oliver Zendel, Christian Theobalt, Alan Yuille, Adam Kortylewski
\IEEEcompsocitemizethanks{
\IEEEcompsocthanksitem B.Z. is with the University of Edinburgh.\\
J.W., W.M., and A.Y. are with Johns Hopkins University.\\
C.T., and A.K. are with MPII.\\\
A.J. and A.K. are with University of Freiburg.\\
Siwei Y. is with the University of California Santa Cruz.\\
Shaozuo Y. is with the Chinese University of Hong Kong.\\
Email: bingchen.zhao@ed.ac.uk.
}
\thanks{Manuscript received April 19, 2005}
}


\IEEEtitleabstractindextext{
\begin{abstract}
Enhancing the robustness of vision algorithms in real-world scenarios is challenging. One reason is that existing robustness benchmarks are limited, as they either rely on synthetic data or ignore the effects of individual nuisance factors. We introduce \OURS, a benchmark dataset that includes out-of-distribution examples of 10 object categories in terms of pose, shape, texture, context and the weather conditions, and enables benchmarking of models for image classification, object detection, and 3D pose estimation. 
In addition to this novel dataset, we contribute extensive experiments using popular baseline methods, which reveal that: 1) Some nuisance factors have a much stronger negative effect on the performance compared to others, also depending on the vision task. 2) Current approaches to enhance robustness have only marginal effects, and can even reduce robustness. 3) We do not observe significant differences between convolutional and transformer architectures. We believe our dataset provides a rich test bed to study robustness and will help push forward research in this area.
Our dataset can be accessed from \href{here}{https://bzhao.me/OOD-CV/}.
\end{abstract}

}

\maketitle

\IEEEpeerreviewmaketitle

\ifCLASSOPTIONcaptionsoff
  \newpage
\fi

\IEEEraisesectionheading{\section{Introduction}\label{sec:intro}}

\begin{figure*}
\centering
\includegraphics[width=\linewidth]{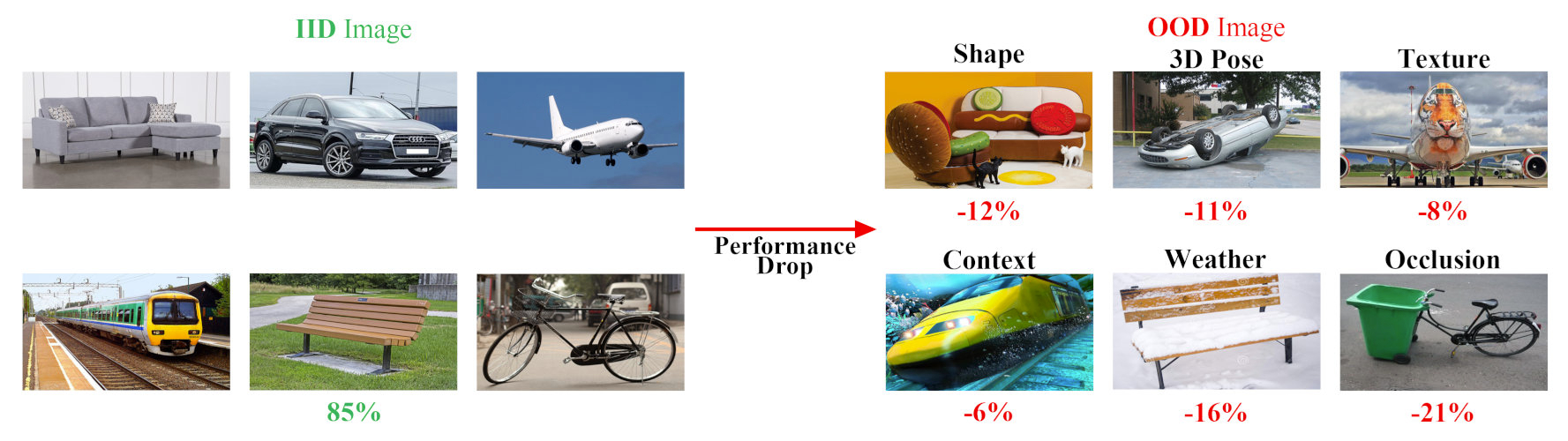}
\caption{{\color{black}Computer vision models are not robust to real-world distribution shifts at test time. For example, ResNet50 achieves about $85\%$ accuracy when tested on images that are similarly distributed as the training data (IID). However, the performance deteriorates significantly when individual nuisance factors in the test images break the IID assumption. Our benchmark makes it possible, for the first time, to study the robustness of image classification, object detection, 3D pose estimation and 6D pose estimation to OOD shifts in individual nuisance variables, including OOD changes in shape, pose, texture, context, weather and partial occlusion.}}
\label{fig:intro}
\end{figure*}

\IEEEPARstart{D}{eep} learning sparked a tremendous increase in the performance of computer vision systems over the past decade, under the implicit assumption that training and test data are drawn independently and identically distributed (IID). However, Deep Neural Networks (DNNs) are still far from reaching human-level performance at visual recognition tasks in real-world environments. The most important limitation of DNNs is that they fail to give reliable predictions in unseen or adverse viewing conditions, which would not fool a human observer, such as when objects have an unusual pose, texture, shape, or when objects occur in an unusual context or in challenging weather conditions (Figure \ref{fig:intro}). The lack of robustness of DNNs in such out-of-distribution (OOD) scenarios is generally acknowledged as one of the core open problems of deep learning, for example by the Turing award winners Yoshua Bengio, Geoffrey Hinton, and Yann LeCun \cite{bengio2021deep}. However, the problem largely remains unsolved.

One reason for the limited progress in OOD generalization of DNNs is the lack of benchmark datasets that are specifically designed to measure OOD robustness.
Historically, datasets have been pivotal for advancement of the computer vision field, e.g. in image classification \cite{deng2009imagenet}, segmentation \cite{lin2014microsoft,everingham2015pascal}, pose estimation \cite{xiang2014beyond,tremblay2018falling,xiang2017posecnn}, and part detection \cite{chen2014detect}.
However, benchmarks for OOD robustness have important limitations, which restrict their usefulness for real-world scenarios. Limitations of OOD benchmarks can be categorized into three types:
Some works measure robustness by training models on one dataset and testing them on another dataset without fine-tuning \cite{rovi,ye2021ood,hendrycks2019robustness,hendrycks2021nae}. 
This cross-dataset performance is only a very coarse measure of robustness and ignores the effects of OOD changes to individual nuisance factors such as the object texture, shape or context.
Other approaches artificially generate corruptions of individual nuisance factors, such as weather \cite{michaelis2019dragon}, synthetic noise \cite{hendrycks2019robustness} or partial occlusion \cite{wang2020robust}.
However, some nuisance factors are difficult to simulate, such as changes in the object shape or 3D pose. Moreover, artificial corruptions, like synthetic noise, only have limited generalization ability to real-world scenarios.
The third type of approach obtains detailed annotation of nuisance variables by recording objects in fully controlled environments, such as in a laboratory \cite{ilab} or using synthetic data \cite{kortylewski2018empirically}. But such controlled recording can only be done for limited amount of objects and it remains unclear if the conclusions made transfer to real-world scenarios. 

In this work, we introduce \OURS, a dataset for benchmarking OOD robustness on real images with annotations of individual nuisance variables and labels for several vision tasks. Specifically, the training and IID testing set in \OURS consists of $10$ rigid object categories from the PASCAL VOC 2012 \cite{pascal-voc-2012} and ImageNet \cite{deng2009imagenet} datasets, and the respective labels for image classification, object detection, as well as the 3D pose annotation from the PASCAL3D+ dataset \cite{xiang2014beyond}.
Our main contribution is the collection and annotation of a comprehensive out-of-distribution test set consisting of images that vary w.r.t. the training data in PASCAL3D+ in terms individual nuisance variables, i.e. images of objects with an unseen shape, texture, 3D pose, context or weather (\cref{fig:intro}). 
Importantly, we carefully select the data such that each of our OOD data samples only varies w.r.t. one nuisance variable, while the other variables are similar as observed in the training data. 
We annotate data with class labels, object bounding boxes and 3D object poses, resulting in a total dataset collection and annotation effort more than 650 hours.
Our \OURS dataset, for the first time, enables studying the influence of individual nuisances on the OOD performance of vision models. 
In addition to the dataset, we contribute an extensive experimental evaluation of popular baseline methods for each vision task and make several interesting observations, most importantly:
1) Some nuisance factors have a much stronger negative effect on the model performance compared to others. Moreover, the negative effect of a nuisance depends on the downstream vision task, because different tasks rely on different visual cues.
2) Current approaches to enhance robustness using strong data augmentation have only marginal effects in real-world OOD scenarios, and sometimes even reduce the OOD performance. Instead, some results suggest that architectures with 3D object representations have an enhanced robustness to OOD shifts in the object shape and 3D pose.
3) We do not observe any significant differences between convolutional and transformer architectures in terms of OOD robustness.
We believe our dataset provides a rich testbed to benchmark and discuss novel approaches to OOD robustness in real-world scenarios and we expect the benchmark to play a pivotal role in driving the future of research on robust computer vision.
%

{\color{black}
Finally, we note that this article extends the conference paper \cite{zhao2022ood} in multiple ways: (1) We largely increase the dataset size by $250\%$ and annotate $6557$ additional images, corresponding to $1600$ additional hours of annotation. (2) We add partial occlusion as additional nuisance variable, complementing the existing nuisances: context, texture, shape, pose and weather. (3) We significantly extend the experimental section by re-running all experiments for classification, detection and 3D pose estimation on the newly collected extended dataset. Moreover, we additionally study the tasks of 6D Pose estimation (Section \ref{sec:exp:occ}), and 3D-aware classification (Section \ref{sec:exp:3dclass}) where algorithms need to jointly estimate the 3D object pose and the class label. (4) We add an in-depth discussion of the results (Section \ref{sec:summary_disc}) including new illustrations where applicable.
}
\section{Related works}



\noindent \textbf{Robustness benchmark on synthetic images.}~~
There has been a lot of recent work on utilizing synthetic images to test the robustness of neural networks~\cite{kurakin2016adversarial,hendrycks2019robustness,michaelis2019dragon}.
For example, ImageNet-C~\cite{hendrycks2019robustness} evaluates the performance of neural networks on images with synthetic noises such as JPEG compression, motion-blur and Gaussian noise by perturbing the standard ImageNet~\cite{deng2009imagenet} test set with these noises. 
\cite{michaelis2019dragon} extends this idea of perturbing images with synthetic noises to the task of object detection by adding these noises on COCO~\cite{lin2014microsoft} and Pascal-VOC~\cite{everingham2015pascal} test sets.
Besides perturbation from image processing pipelines, there are also work~\cite{geirhos2018} benchmarks the shape and texture bias of DNNs using images with artificially overwritten textures.
Using style-transfer~\cite{gatys2016image} as augmentation~\cite{geirhos2018} or using a linear combination between strongly augmented images and the original images~\cite{hendrycks2019augmix} have been shown as effective ways of improving the robustness against these synthetic image noises or texture changes.
However, these benchmarks are limited in a way that synthetic image perturbations are not able to mimic real-world 3-dimensional nuisances such as novel shape or novel pose of objects.
Our experiments in~\cref{sec:exp} also show that style-transfer~\cite{gatys2016image} and strong augmentation~\cite{hendrycks2019augmix} does not help with shape and pose changes.
In addition, these benchmarks are limited to single tasks, for example, ImageNet-C~\cite{hendrycks2019robustness} only evaluates the robustness on image classification, COCO-C~\cite{michaelis2019dragon} only evaluates on the tasks of object detection.
DomainBed~\cite{gulrajani2020search} also benchmarks algorithm on OOD domain generalization on the task of classification.
In our work, we evaluate the robustness on real world images, while also evaluate the robustness across different tasks including image classification, object detection, and pose estimation.

\begin{figure*}
    \centering
    \includegraphics[width=0.95\textwidth]{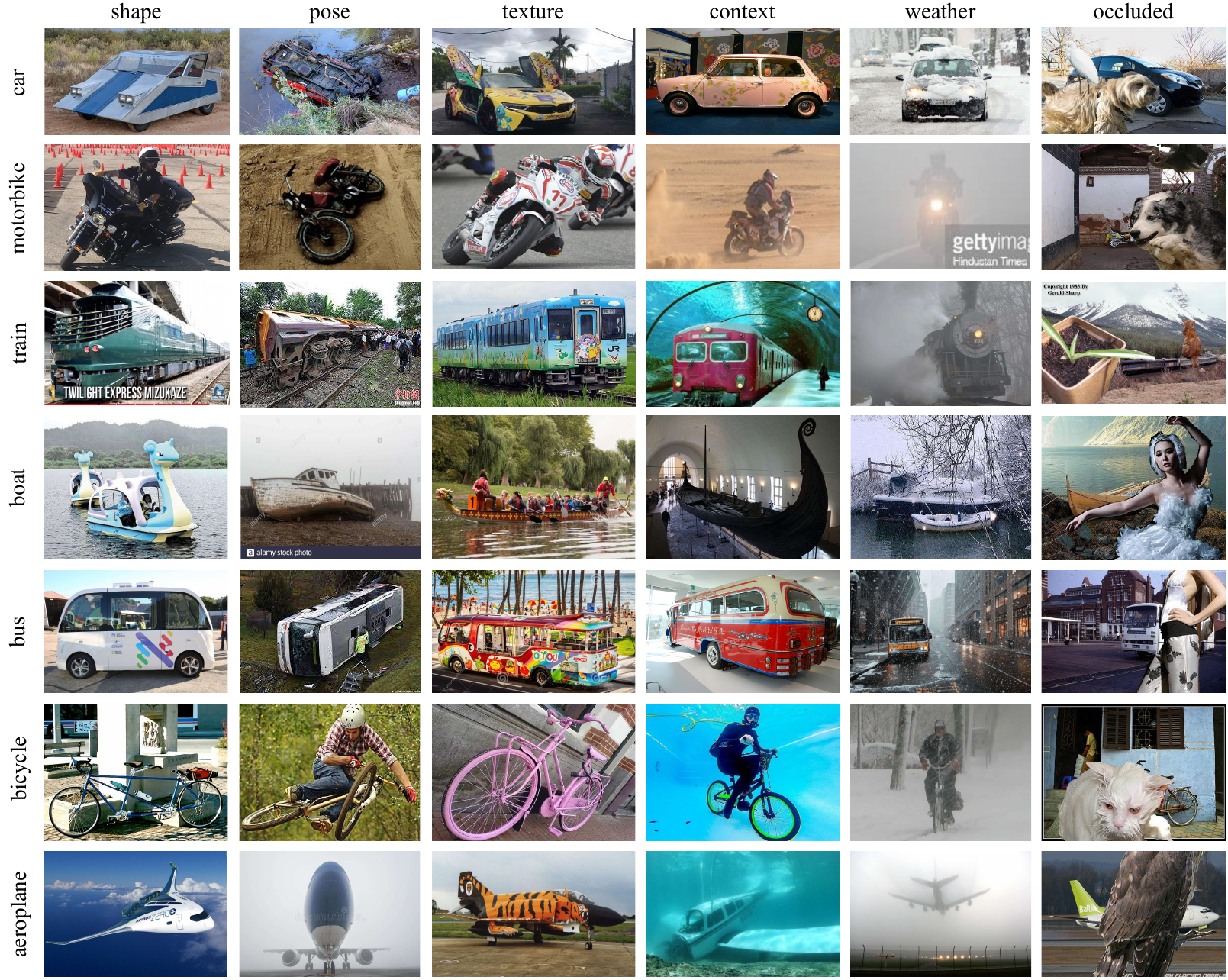}
    \caption{{\color{black}Examples from our dataset with OOD variations of individual nuisance factors including the object shape, pose, texture, context, weather, and occluded conditions.}
    }
    \label{fig:data_demo}
\end{figure*}

\noindent \textbf{Robustness benchmark on real world images.}~~
Distribution shift in real-world images are more than just synthetic noises, many recent works~\cite{recht2019imagenet,hendrycks2021nae,hendrycks2019robustness} focus on collecting real-world images to benchmark robustness of DNN performances.
ImageNet-V2~\cite{recht2019imagenet} created a new test set for ImageNet~\cite{deng2009imagenet} by downloading images from Flickr, and found this new test set causes the model performance to degrade, showing that the distribution shift in the real images has an important influence on DNN models.
By leveraging an adversarial filtration technique that filtered out all images that a fixed ResNet-50~\cite{he2015deep} model can correctly classifies, ImageNet-A~\cite{hendrycks2021nae} collected a new test set and shows that these adversarially filtered images can transfer across other architectures and cause the performance to drop by a large margin.
Although ImageNet-A~\cite{hendrycks2021nae} shows the importance of evaluating the robustness on real-world images, but cannot isolate the nuisance factor. 
The Wilddash 2 segmenation dataset and benchmark \cite{Zendel_2022_CVPR} focuses on difficult road scenes. Their benchmark dataset is grouped by ten identified nuisances (e.g. interior reflections, unusual road coverage, overexposure) called \textit{visual hazards} based on results of a risk analysis method\cite{Zendel2017}. The online benchmark service calculates performance drops for each nuiance by comparing average perfomance from IID and nuisance subsets.
Most recently, ImageNet-R~\cite{hendrycks2021many} collected four OOD testing benchmarks by collecting images with distribution shifts in texture, geo-location, camera parameters, and blur respectively, and shows that not one single technique can improve the model performance across all the nuisance factors.
There are also benchmarks to test how well a model can learn invariant features from unbalanced datasets~\cite{tang2022invariant}.
And benchmarks composed of many real-world shifts~\cite{koh2021wilds}.
We introduce a robustness benchmark that is complementary to prior datasets, by disentangling individual OOD nuisance factors that correspond to semantic aspects of an image, such as the object texture and shape, the context object, and the weather conditions. Due to rich annotation of our data, our benchmark also enables studying OOD robustness for various vision tasks.

\noindent \textbf{Techniques for improving robustness.}~~
To close the gap between the performance of vision models on datasets and the performance in the real-world, many techniques has been proposed~\cite{Mohseni2021PracticalML}. 
These techniques for improving robustness can be roughly categorized into two types: data augmentation and architectural changes.
Adversarial training by adding the worst case perturbation to images at training-time~\cite{wong2020fast}, using stronger data augmentation~\cite{cubuk2018autoaugment,wang2021augmax}, image mixtures~\cite{hendrycks2019augmix,yun2019cutmix,erichson2022noisymix}, and image stylizations~\cite{geirhos2018} during training, or augmenting in the feature space~\cite{hendrycks2021many} are all possible methods for data augmentation.
These data augmentation methods have been proven to be effective for synthetic perturbed images~\cite{hendrycks2019augmix,geirhos2018}.
Architectural changes are another way to improve the robustness by adding additional inductive biases into the model.
\cite{xie2019feature} proposed to perform de-noise to the feature representation for a better adversarial robustness. 
Analysis-by-synthesis appoaches~\cite{wang2021nemo,kortylewski2021compositional} can handle scenarios like occlusion by leveraging a generative object model and through top-down feedback \cite{xiao2020tdmpnet}.
Transformers are a newly emerged architecture for computer vision~\cite{dosovitskiy2020image,liu2021swin,Shao_2021_WACV}, and there are works showing that transformers may have a better robustness than CNNs~\cite{bhojanapalli2021understanding,mahmood2021robustness}, although our experiments suggest that this is not the case.
Object-centric representations~\cite{locatello2020object,wen2022selfsupervised} have also been show to improve robustness.
Self-supervised learned representations also show improvement on OOD examples~\cite{hendrycks2019selfsupervised,zhao2020distilling,zhu2021improving,cui2022discriminabilitytransferability}
Our benchmark enables the comprehensive evaluation of such techniques to improve the robustness of vision models on realistic data, w.r.t. individual nuisances and vision tasks. We find that current approaches to enhance robustness have only marginal effects, and can even reduce robustness, thus highlighting the need for an enhanced effort in this research direction. 

\begin{figure*}
    \centering
    \includegraphics[width=0.95\textwidth]{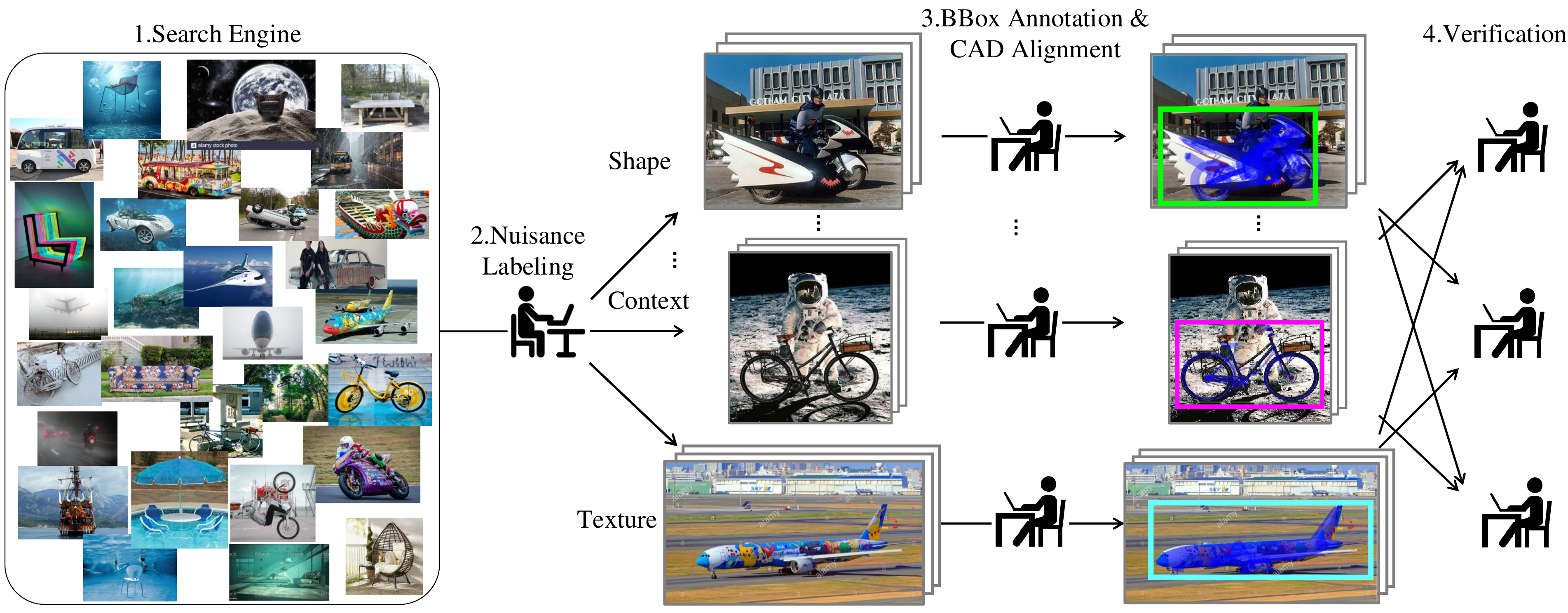}
    \caption{Data is collected from the internet using a predefined set of search keywords. All images are manually filtered to remove those lacking OOD nuisances or having multiple nuisances. After collecting and splitting the data into different collections with different nuisance, we label images with object bounding boxes and align a CAD model to estimate the 3D pose of the object. The CAD models are overlaid on the images in blue. After each image annotation has been verified by at least two other annotators, we include it in our final dataset.}
    \label{fig:data_collection}
\end{figure*}

\section{Dataset collection}\label{sec:collection}

\begin{table*}
\centering
\caption{\label{tab:stats_per_cls} 
{\color{black}The statistics of the classification subset in our dataset, i.e. the number of images having objects from each category with individual nuisances. There are 17649 images in total in our test set. Among them, 13762 images are OOD images (11130 more images than the OOD images in the preliminary version~\cite{zhao2022ood}). To benchmark the IID performance, we also obtain 3887 images from the PASCAL3D+ dataset.}}
{\color{black}
\begin{tabular}{lrrrrrrrr}
\toprule
\#Instances      & Context & Occlusion & Pose & Shape & Texture & Weather & IID & Total\\
\midrule
aeroplane        & 176     & 746     & 215   & 464  & 101     & 262       & 123  & 2087 \\
bicycle          & 244     & 446     & 207   & 253  & 72      & 154       & 262  & 1638 \\
boat             & 230     & 683     & 171   & 93   & 106     & 183       & 571  & 2037 \\
bus              & 289     & 315     & 131   & 165  & 149     & 90        & 623  & 1762 \\
car              & 120     & 1804    & 242   & 151  & 58      & 191       & 321  & 2887 \\
chair            & 251     & 251     & 103   & 594  & 302     & 54        & 315  & 1870 \\
diningtable      & 29      & 724     & 41    & 164  & 294     & 15        & 238  & 1505 \\
motorbike        & 83      & 396     & 193   & 111  & 71      & 154       & 279  & 1287 \\
sofa             & 138     & 75      & 49    & 153  & 111     & 12        & 318  & 856 \\
train            & 83      & 339     & 74    & 102  & 118     & 167       & 837  & 1720 \\
\midrule
Total            & 1643    & 5779    & 1426  & 2250 & 1382    & 1282      & 3887 & 17649 \\
\bottomrule
\end{tabular}
}
\end{table*}


In this section, we introduce the design of the \OURS benchmark and discuss the data collection process to obtain the OOD images and annotations.

\subsection{What are important nuisance factors?}
\label{sec:ontology}
The goal of the \OURS benchmark is to measure the robustness of vision models to realistic OOD shifts w.r.t. important individual nuisance factors.
To achieve this, we define an ontology of nuisance factors that are relevant in real-world scenarios following related work on robust vision \cite{michaelis2019benchmarking,rosenfeld2018elephant,qiu2016unrealcv,alcorn2019strike,kortylewski2019analyzing,Zendel2017} and taking inspiration from the fact that images are 3D scenes with a hierarchical compositional structure, where each component can vary independently of the other components. 
In particular, we identify \new{six} important nuisance factors that vary strongly in real-world scenarios: object shape, 3D pose, texture appearance, surrounding context, weather conditions, \new{and occlusion}.
These nuisance factors can be annotated by a human observer with reasonable effort, while capturing a large amount of the variability in real-world images.
Notably, each nuisance can vary independently from the other nuisance factors, which will enable us to benchmark the OOD effect of each nuisance individually. 

\subsection{Collecting images}
OOD data can only be defined w.r.t. some reference distribution of training data.  
For our dataset, the reference training data is based on the PASCAL3D+~\cite{xiang_wacv14} dataset which is composed of images from Pascal-VOC~\cite{everingham2015pascal} and ImageNet~\cite{deng2009imagenet} datasets, and contains annotations of the object class, bounding box and 3D pose. 
Our goal is to collect images where only one nuisance factor is OOD w.r.t. training data, while other factors are similar as in training data.

To collect data with OOD nuisance factors, we search the internet using a curated set of search keywords that are combinations of the object class from the PASCAL3D+ dataset and attribute words that may retrieve images with OOD attributes, e.g. "car+hotdog" or "motorbike+batman", a comprehensive list of our search keywords used can be found in the appendix. \new{Besides using Google as the search engine as in~\cite{zhao2022ood}, we also add some Chinese keywords and search them in Baidu, the most dominant Chinese search engine, to increase the number of images in our dataset.}
Note that we only use $10$ object categories from PASCAL3D+, 
as we could not find sufficient OOD test samples for all nuisances for the categories "bottle" and "television".
We manually filtered images with multiple nuisances and put an effort in retaining images that significantly vary in terms of one nuisance only.
Following this approach, we collect  \new{1643, 1382, 2250, and 1282 instances} with OOD nuisances in terms of context, texture,  shape, and weather, \new{respectively}. 

We leverage the shape and pose annotations from PASCAL3D+ to create OOD dataset splits regarding 3D pose and shape. These allow us to split the dataset such that 3D pose and shape of training and testing set do not overlap. 
We augment these OOD splits in pose and shape with additional data that we collect from the internet. \new{In this way, we collect  $1426$ and $2250$ instances with OOD nuisances in 3D pose and shape respectively}. 

\new{Different from the preliminary version~\cite{zhao2022ood} of our dataset, we add \textit{occlusion} as a new nuisance factor. As in~\cite{wang2020robust}, we use animals, plants, and humans cropped from MS-COCO dataset~\cite{lin2014microsoft} as the occluders. To mimic the real-world occlusions, we superimpose the occluders not only inside the bounding box of the objects (40-60\% of the object area occluded) but also on the background (20-40\% of the context area occluded). Example images are shown in Figure \ref{fig:data_demo}. }

\new{To ensure that the test data is really OOD, three annotators went through all training data from PASCAL3D+ and filtered out images from the training set that were too similar to OOD test data.}


\subsection{Detailed statistics}


\new{Statistics of our dataset are shown in Table \ref{tab:stats_per_cls}. }
Overall, the \OURS benchmark is an image collection with a total of \new{$26181$} images composed from PASCAL3D+ and the internet where \new{$18198$} images are from PASCAL3D+ for both training and testing \new{(IID and occluded data)} and \new{$7983$} images are collected and annotated by us testing OOD performance \new{on 5 nuisances (context, pose, shape, texture, and weather)}. 
\new{On average we have \new{$229$ instances} per nuisance and object class which is \new{higher than} other datasets, e.g. ImageNet-C with an average of $50$ images.}
To enable us to benchmark OOD robustness, the nuisance factors and vision tasks were annotated as discussed in the next section.
\new{Note that due to the different nature of image classification, object detection, and pose estimation tasks and the difficulty of annotating the images, the number of images that are used for different tasks is different, we provide a detailed statistics of the three tasks in the supplementary.}

\subsection{Data annotation}
\label{sec:annotation}
A schematic illustration of the annotation process is shown in ~\cref{fig:data_collection}.
After collecting the images from the internet, we first classify the images according to the OOD nuisance factor following the ontology discussed in~\cref{sec:ontology}. Subsequently, we annotate the images to enable benchmarking of a variety of vision tasks. In particular, we annotate the object class, 2D bounding box, and 3D object pose. 
Note that we include the 3D pose, despite the large additional annotation effort compared to class labels and 2D bounding boxes, because we believe that extracting 3D information from images is an important computer vision task. 

The annotation of the bounding boxes follows the coco format~\cite{lin2014microsoft}. We used a web-based annotation tool~\footnote{https://github.com/jsbroks/coco-annotator} that enables the data annotation with multiple annotators in parallel. \new{Besides, different from \cite{zhao2022ood}, we collect additional bounding box annotations for the newly added images by using Amazon Mechanical Turk (AMT). For each image, we have 3 experienced annotators (each has at least 5000 approved annotations and the approval rate should be higher than 98\%).}

The 3D pose annotation mainly follows the pipeline of PASCAL3D+~\cite{xiang_wacv14} and we use a slightly modified annotation tool from the one used in the PASCAL3D+ toolkit~\footnote{https://cvgl.stanford.edu/projects/pascal3d.html}. Specifically, to annotate the 3D pose each annotator selects a CAD model from the ones provided in PASCAL3D+, which best resembles the object in the input image. 
Subsequently, the annotator labels several keypoints to align the 6D pose of the CAD model to the object in the input image.
After we have obtained annotations for the images, we count the distribution of number of images in each category and for categories with fewer images than average, we continue to collect additional images from the internet for the minority categories.
Following this annotation process, we collected labels for all \new{$7983$} images covering \new{context, pose, shape, texture, and weather nuisances}.
Finally, the annotations produced by every annotator are verified by at least two other annotators to ensure the annotation is correct.
We have a total of $5$ annotators, and it took about $15$ minutes per image, resulting in around \new{$2000$} hours of annotation effort~\footnote{\new{We would like to thank the researchers at Austrian Institute Of Technology (AIT, https://www.ait.ac.at) for their help in annotating part of the data.}}.
\begin{table*}[h]
\small
\centering
\caption{\label{tab:individual_nuisance_on_different_tasks} Robustness to individual nuisances of popular vision models for different vision tasks. 
We report the performance on i.i.d. test data and OOD shifts in the object shape, 3D pose, texture, context and weather.
Note that image classification models are most affected by OOD shifts in the weather, while detection and pose estimation models are mostly affected by OOD shifts in context and shape, suggesting that vision models for different tasks rely on different visual cues.}
\resizebox{\textwidth}{!}{

{\color{black}
\begin{tabular}{c|lccccccc}
\toprule
Task  &                  & i.i.d             & shape             & pose              & texture          & context     & occlusion      & weather   \\ 
\midrule
\multirow{2}{*}{\begin{tabular}{c}Image\\Classification\end{tabular}} 
& ResNet-50              & 83.9\%$\pm$0.2\% & 68.7\%$\pm$0.3\% & 76.1\%$\pm$0.1\% & 67.6\%$\pm$0.4\% & 65.1\%$\pm$0.3\%  & 58.5\%$\pm$0.2\% & 71.5\%$\pm$0.3\%  \\
 & MbNetv3-L             & 79.5\%$\pm$0.3\% & 64.3\%$\pm$0.2\% & 71.3\%$\pm$0.6\% & 62.3\%$\pm$0.3\% & 60.2\%$\pm$0.3\% &  53.2\%$\pm$0.4\%  & 64.7\%$\pm$0.5\%   \\
\midrule
\multirow{2}{*}{\begin{tabular}{c}Object\\Detection\end{tabular}}
 & Faster-RCNN           & 40.6\%$\pm$0.3\% & 34.9\%$\pm$0.3\% & 30.4\%$\pm$0.3\% & 34.6\%$\pm$0.5\% & 27.0\%$\pm$0.2\% & 10.0\%$\pm$0.6\% & 28.3\%$\pm$0.4\% \\ 
 & RetinaNet             & 43.9\%$\pm$0.4\% & 39.2\%$\pm$0.3\% & 33.4\%$\pm$0.5\% & 37.4\%$\pm$0.2\% & 27.8\%$\pm$0.6\% & 16.8\%$\pm$0.3\% & 31.3\%$\pm$0.3\% \\ 
\midrule
\multirow{2}{*}{\begin{tabular}{c}3D Pose\\Estimation\end{tabular}}
 & Res50-Specific         & 62.4\%$\pm$2.4\% & 43.5\%$\pm$2.5\% & 45.2\%$\pm$2.8\% & 51.4\%$\pm$1.8\% & 50.8\%$\pm$1.9\% &  
41.6\%$\pm$2.1\%   & 49.5\%$\pm$2.1\% \\ 
 & NeMo                  & 66.7\%$\pm$2.3\% & 51.7\%$\pm$2.3\% & 56.9\%$\pm$2.7\% & 52.6\%$\pm$2.0\% & 51.3\%$\pm$1.5\% & 62.2\%$\pm$2.7\% & 49.8\%$\pm$2.0\% \\ 
\bottomrule
\end{tabular}}
}

\end{table*}

\noindent \textbf{Dataset splits.} 
To benchmark the IID performance, we split the \new{$12419$} images that we retained from the PASCAL3D+ dataset into \new{$8532$} training images and \new{$3887$} test images.
The OOD dataset splits for the nuisances "texture", "context", and "weather" can be directly used from our collected data.
As the Pascal3D+ data is highly variable in terms of 3D pose and shape, we create OOD splits w.r.t. the nuisances "pose" and "shape" by biasing the training data using the pose and shape annotations, such that the training and test set have no overlap in terms of shape and pose variations. These initial OOD splits are 
further enhanced using the data we collected from the internet. 
The dataset and detailed documentation of the dataset splits is available online\footnote{http://ood-cv.org/, Also see the supplementary material.}.


\section{Experiments}
\label{sec:exp}


We test the robustness of vision models w.r.t. out-of-distribution shifts of individual nuisance factors in~\cref{sec:individual_nuisance} and evaluate popular methods for enhancing the model robustness of vision models using data augmentation techniques (\cref{sec:data_aug}) and changes to the model architecture (\cref{sec:model_architecture}).
Finally, we study the effect when multiple nuisance factors are subject to OOD shiftsin~\cref{sec:combined_nuisance} and give a comprehensive discussion of our results in~\cref{sec:summary_disc}.

\subsection{Experimental Setup}
Our \OURS dataset enables benchmark vision models for three popular vision tasks: image classification, object detection, and 3D pose estimation.
We study robustness of popular methods for each task w.r.t. OOD shifts in six nuisance factors: object shape, 3D pose, object texture, background context, novel occlusion and weather conditions.
We use the standard evaluation metrics of mAP and Acc@$\frac{\pi}{6}$ for object detection and 3D pose estimation respectively.
For image classification, we crop the objects in the images based on their bounding boxes to create object-centric images, and use the commonly used Top-1 Accuracy to evaluate the performance of classifiers.
In all our experiments, we control variables such as the number of model parameters, model architecture, and training schedules to be comparable and only modify those variables we wish to study. The models for image classification are pre-trained on ImageNet \cite{deng2009imagenet} and fine-tuned on our benchmark. As datasets for a large-scale pre-training are not available for 3D pose estimation, we randomly initialize the pose estimation models and directly train them on the \OURS training split.

\textbf{Implementation Details}
In the following, we discuss the detailed training settings for vision models, data splits and techniques for improving the robustness in our experiments.

\textit{Image Classification.} 
For the experiments of image classification on \OURS datasets, we tested three network architectures, namely, MobileNetV3-Large~\cite{howard2019searching_mbv3}, ResNet-50~\cite{he2015deep}, and Swin-T~\cite{liu2021swin}.
We train all three models with the same hyper-parameter to make a fair comparison.
The Batchsize is set to \zbc{256} with a step-decayed learning rate initialized with \zbc{0.03} and then multiplied by \zbc{30,60,90} epochs, we train the network for a total of \zbc{90} epochs on the training set.
The resolution of the input images are 224 by 224 which is also a default value for training networks~\cite{he2015deep}.

We compared the effectiveness of different data augmentation techniques, namely, style transfer~\cite{geirhos2018}, AugMix~\cite{hendrycks2019augmix}, and \zbc{PixMix~\cite{pixmix}}.
For all the experiments using style transfer~\cite{geirhos2018}, we use the code from the original authors~\footnote{https://github.com/rgeirhos/Stylized-ImageNet} to create the style augmented images for training.
For experiments with AugMix~\cite{hendrycks2019augmix}, we adopted a PyTorch-based implementation~\footnote{https://github.com/psh150204/AugMix}. 
For experiments with PixMix~\cite{pixmix}, we adopted the official implementation~\footnote{https://github.com/andyzoujm/pixmix}.

\textit{Object Detection.} We mainly used two frameworks for the task of object detection, namely Faster-RCNN~\cite{ren2015faster} and RetinaNet~\cite{lin2017focal}.
Similarly, we keep all the hyper-parameter the same except for the ones we wish to study.
The experiments are mainly conducted using the detectron2 codebase~\footnote{https://github.com/facebookresearch/detectron2}.
For strong data augmentation techniques that can be used to improve the robustness of vision models, AugMix~\cite{hendrycks2019augmix} is relatively harder to implement than the other on object detection because of the image mixing step, so we only evaluated the performance of style transfer.
The style transfer uses the same images generated for image classification.

We train all the object detection models with 18000 iterations with an initial learning rate of 0.02 and a batchsize of 16, the learning rate is then multiplied by 0.1 at 12000 and 16000 iterations.
We adopted the multi-scale training technique to improve the baseline performance, each input images will be resized to have a short edge of $[480, 512, 544, 576, 608, 640, 672, 704, 736, 768, 800]$, and when testing, the test input image will be resized to have a short edge of 800.
For experiments with Swin-T as the backbone network in the detection framework, we adopted the implementations from the authors of the swin-transformer~\footnote{https://github.com/SwinTransformer/Swin-Transformer-Object-Detection}.

\textit{3D pose estimation.} For 3D pose estimation, we evaluated two types of models, Res50-Specific~\cite{zhou2018starmap} and NeMo~\cite{wang2021nemo}.
We adopted the implementation from the original authors~\footnote{https://github.com/shubhtuls/ViewpointsAndKeypoints}\footnote{https://github.com/Angtian/NeMo}.
When training the pose estimation models, we use a batchsize of 108 and a learning rate of 1e-3.
For the pose estimation model for each category, we train the model for 800 epochs.



\subsection{Robustness to individual nuisances}
\label{sec:individual_nuisance}
The \OURS benchmarks enables, for the first time, to study the influence of OOD shifts in individual nuisance factors on tasks of classification, detection and pose estimation.
We first study the robustness of one representative methods for each task. 
In~\cref{tab:individual_nuisance_on_different_tasks}, we report the test performance on a test set with IID data, as well as the performance under OOD shifts to all six nuisance factors that are annotated in the \OURS benchmark.
We observe that for image classification, the performance of the classic ResNet50 architecture~\cite{he2015deep} drops significantly for every OOD shift in the data. The largest drop is observed under OOD shifts in the occlusion conditions ($-15.4\%$), while the performance drop for OOD pose is only $-7.8\%$. 
\zbc{Occlusion causes the biggest drop in both classification and detection.}
The results suggest that the model 
On the contrary, for object detection the performance of a Faster-RCNN~\cite{ren2015faster} model drops the most under OOD context ($-37\%$ mAP), showing that detection models rely strongly on contextual cues. 
While the performance of the detection model also decreases significantly across all OOD shifts, the appearance-based shifts like texture, context and weather have a stronger influence compared to OOD shifts in the shape and pose of the object. 
For the task of 3D pose estimation, we study a ResNet50-Specific~\cite{zhou2018starmap} model, which is a common pose estimation baseline that treats pose estimation as a classification problem (discretizing the pose space and then classifying an image into one of the pose bins).
We observe that the performance for 3D pose estimation drops significantly, across all nuisance variables and most prominently for OOD shifts in the shape and pose.

\begin{figure*}
     \centering
     \begin{subfigure}[b]{0.3\textwidth}
         \centering
         \includegraphics[width=\textwidth]{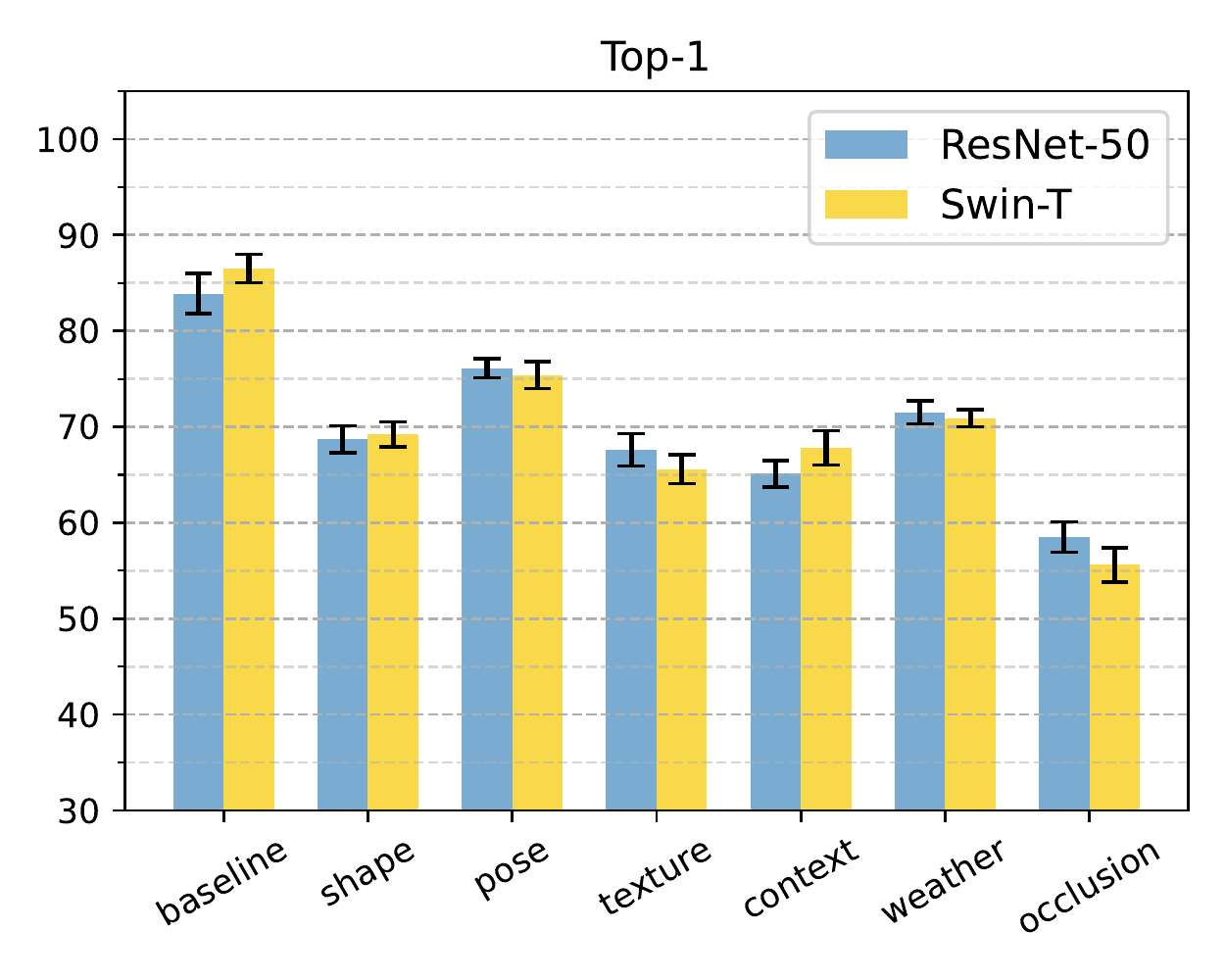}
         \caption{\color{black}Image Classification}
         \label{fig:cnn_vs_transformer_classification}
     \end{subfigure}
     \begin{subfigure}[b]{0.3\textwidth}
         \centering
         \includegraphics[width=\textwidth]{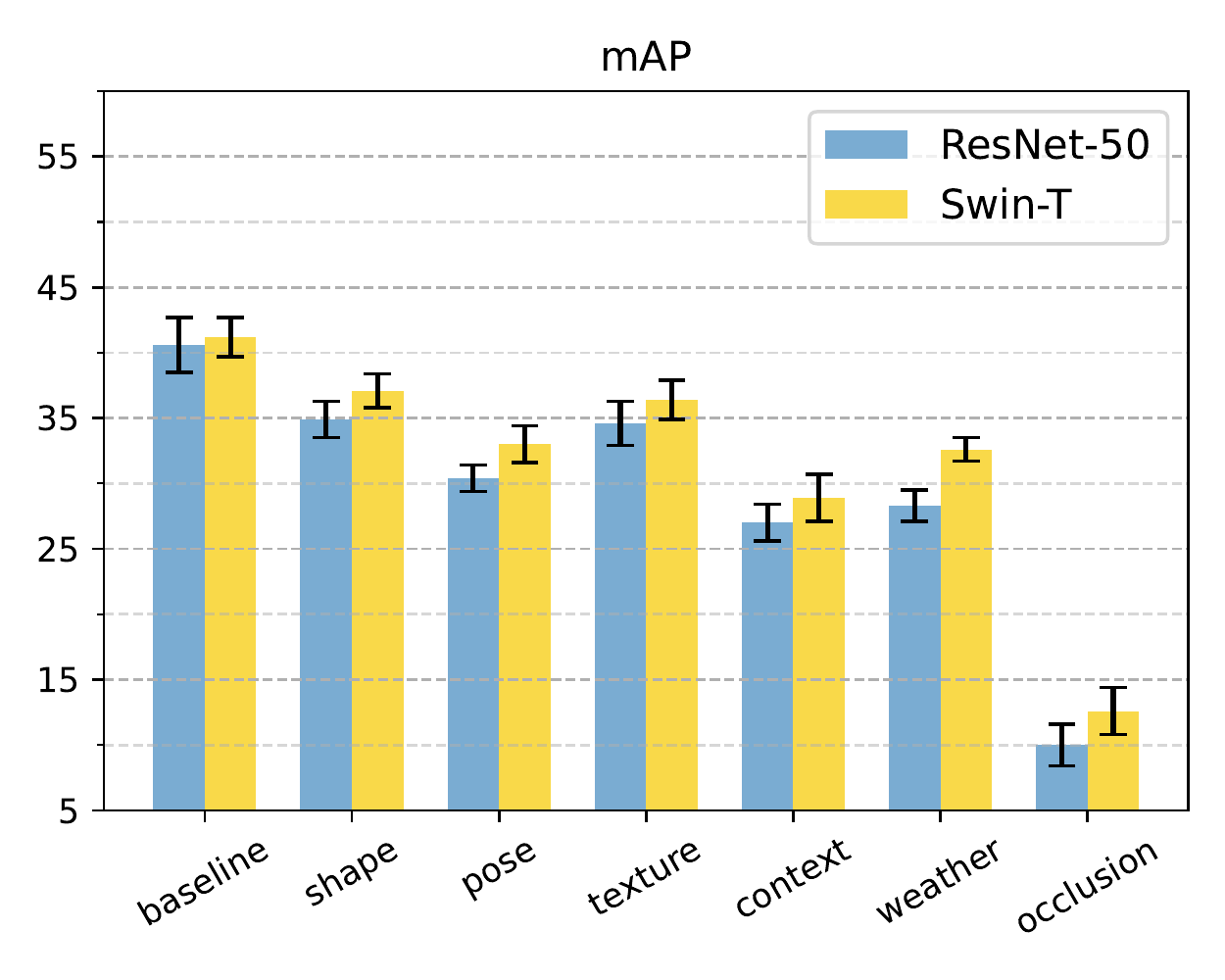}
         \caption{\color{black}Object Detection}
         \label{fig:cnn_vs_transformer_detection}
     \end{subfigure}
     \begin{subfigure}[b]{0.3\textwidth}
         \centering
         \includegraphics[width=\textwidth]{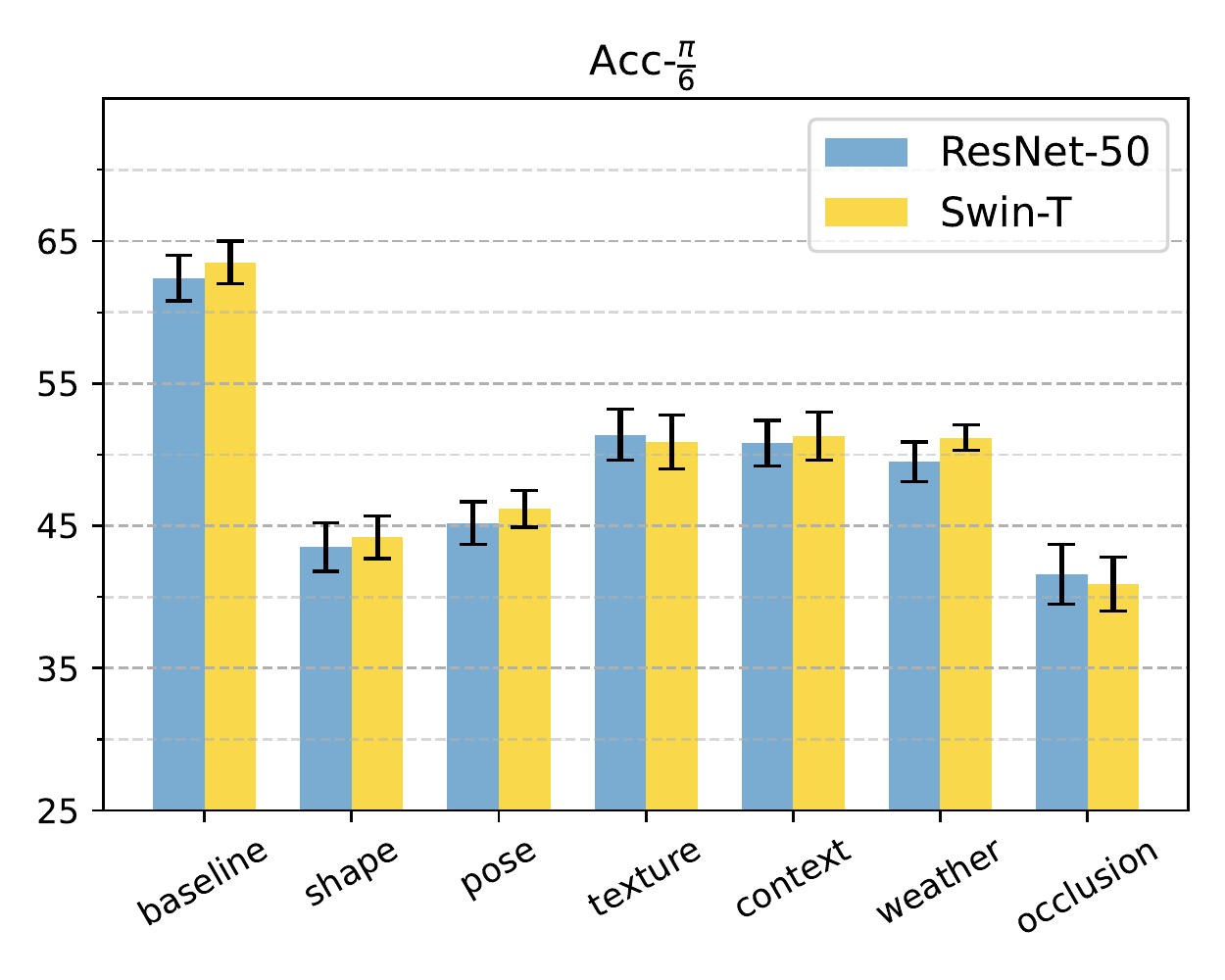}
         \caption{\color{black}3D Pose Estimation}
         \label{fig:cnn_vs_transformer_estimation}
     \end{subfigure}
        \caption{Performance of CNN and Transformer on our benchmark. Transformers have a higher in-domain performance, but CNNs and transformers degrades mostly the same on OOD testing examples. 
        }
        \label{fig:cnn_vit}
        
\end{figure*}

In summary, our experimental results show that \textbf{OOD nuisances have different effect on vision models for different visual tasks}. 
This suggests OOD robustness should not be simply treated as a domain transfer problem between datasets, but instead it is important to study the effects of individual nuisance factors. 
Moreover, OOD robustness might require different approaches for each vision tasks, as we observe clear differences in the effect of OOD shifts in individual nuisance factors between vision tasks.

\begin{table*}
\small
\caption{\zbc{Effect of data augmentation techniques on OOD robustness for three vision tasks. We report the performance of one baseline model for each task, as well as the same model trained with different augmentation techniques: Stylizing, AugMix\cite{hendrycks2019augmix} and PixMix\cite{pixmix}.
We evaluate all models on i.i.d. test data and OOD shifts in the object shape, 3D pose, texture, context and weather. 
Strong data augmentation only improves robustness to appearance-based nuisances but even decreases the performance to geometry-based nuisances like shape and 3D pose.}}
\begin{subtable}{\textwidth}
\centering

{\color{black}
\begin{tabular}{lccccccc}
\toprule
top-1                              & i.i.d & shape & pose &  texture & context & weather & occlusion\\ 
\midrule
ResNet-50                          & 83.9\% & 68.7\% & 76.1\% & 67.6\% & 65.1\% & 71.5\% & 58.5\% \\
Style Transfer                     & 81.2\% & 65.6\% & 75.6\% & 68.3\% & 63.4\% & 69.8\% & 56.7\% \\ 
AugMix                             & 84.5\% & 68.9\% & 76.2\% & 67.1\% & 66.8\% & 74.5\% & 59.7\% \\
PixMix                             & 85.3\% & 69.0\% & 76.0\% & 67.9\% & 66.9\% & 75.6\% & 59.8\% \\
\bottomrule
\end{tabular}
}
\caption{Top-1 accuracy results on image classification}
\label{tab:data_aug_cls}
\end{subtable}

\begin{subtable}{\textwidth}
\centering
{\color{black}
\begin{tabular}{lccccccc}
\toprule
mAP                            & i.i.d   & shape  & pose   & texture & context & weather & occlusion\\ 
\midrule
Faster-RCNN                          & 40.6\% & 34.9\% & 30.4\% & 34.6\% & 27.0\% & 28.3\% & 10.0\% \\
Style Transfer                    & 41.9\% & 36.4\% & 31.6\% & 38.0\% & 31.1\% & 33.4\%& 14.0\%  \\
PixMix                            & 42.0\% & 37.4\% & 32.4\% & 38.0\% & 31.3\% & 34.7\%& 18.5\%  \\ 
\bottomrule
\end{tabular}}
\caption{mAP results on object detection}
\label{tab:data_aug_det}
\end{subtable}

\begin{subtable}{\textwidth}
\centering
{\color{black}
\begin{tabular}{lccccccc}
\toprule
Acc-$\frac{\pi}{6}$               & i.i.d   & shape  & pose   & texture & context & weather & occlusion\\ 
\midrule
Res50-Spec.                         & 62.4\% & 43.5\% & 45.2\% & 51.4\% & 50.8\% & 49.5\% & 41.6\% \\
Style Transfer                    & 63.1\% & 41.8\% & 44.7\% & 55.8\% & 54.3\% & 53.8\% & 40.5\%\\ 
AugMix                            & 64.8\% & 44.1\% & 44.8\% & 56.7\% & 54.7\% & 55.6\% & 42.1\%\\
\bottomrule
\end{tabular}
}
\caption{Acc-$\frac{\pi}{6}$ results on pose estimation}
\label{tab:data_aug_pose}
\end{subtable}

\label{tab:data_aug}
\end{table*}

\subsection{Data Augmentation for Enhancing Robustness}
\label{sec:data_aug}
\zbc{Add discussion with PixMix.}
Data augmentation techniques have been widely adopted as an effective means of improving the robustness of vision models. Among such data augmentation methods, stylizing images with artistic textures~\cite{geirhos2018}, mixing up the original image with a strongly augmented image (AugMix~\cite{hendrycks2019augmix}), and PixMix~\cite{pixmix} are the most effective methods.
We test these data augmentation methods on \OURS to find out if and how they affect the OOD robustness.
The experimental results are summarized in~\cref{tab:data_aug}.
Overall, AugMix~\cite{hendrycks2019augmix} improves the OOD robustness the most for image classification and pose estimation. While AugMix is not directly applicable to object detection, we observe that strong data augmentation style transfer~\cite{gatys2016image,pixmix} leads to a better improvement compared to adversarial training.
Importantly, these data augmentation methods improve the OOD robustness mostly w.r.t. appearance-based nuisances like texture, context, and weather.
However, in all our experiments \textit{data augmentation slightly reduces the performance} under OOD shape and 3D pose.
We suspect that this happens because data augmentation techniques mostly change appearance-based properties of the image and do not change the geometric properties of the object (i.e. shape and 3D pose).
Similar trends are observed across all three of the tasks we tested, image classification, object detection, and pose estimation.
These results suggest that two categories of nuisances exists, namely \textit{appearance-based} nuisances like novel texture, context, and weather, and \textit{geometric-based} nuisances like novel shape and pose.
We observe that \textbf{data augmentation only improves robustness of appearance-based nuisances but can even decrease the performance w.r.t. geometry-based nuisances}.

\subsection{Effect of Model Architecture on Robustness}
\label{sec:model_architecture}
In this section, we investigate four popular architectural changes that have proven to be useful in real world applications. Paricularly, we evaluate \textit{CNNs vs Transformers}, the \textit{model capacity}, \textit{one stage vs two stage} detectors, and models with \textit{integrated 3D priors}. 
Note that when we change the model architecture we keep other parameters such as number of parameters and capacity the same.

\begin{table*}
\small
\centering
\caption{\label{tab:mbv3_vs_r50} OOD robustness of models with different capacities. While the performance degradation of MobileNetv3-Large (MbNetv3-L) are about the same as those of ResNet-50, training with data augmentation technique has smaller effect on MbNetv3-L due to the limited capacity.}
\setlength\tabcolsep{0.3em}
{\color{black}
\begin{tabular}{lccccccc}
\toprule
                                   & i.i.d   & shape  & pose   & texture   & context  & weather & occlusion  \\ 
\midrule
ResNet-50                          & 83.9\% & 68.7\% & 76.1\% & 67.6\% & 65.1\% & 71.5\% & 58.5\% \\
+AugMix                             & 84.5\% & 68.9\% & 76.2\% & 67.1\% & 66.8\% & 74.5\% & 59.7\% \\
\midrule
MbNetv3-L~\cite{howard2019searching_mbv3}     & 79.5\% & 64.3\% & 71.3\% & 62.3\% & 60.2\% & 64.7\% & 53.2\%   \\
+AugMix~\cite{hendrycks2019augmix}          & 80.9\%  & 64.8\%  & 71.8\% & 61.8\% & 61.3\% & 67.8\% & 54.2\%   \\
\bottomrule
\end{tabular}
}
\end{table*}

\noindent \textit{CNNs vs Transformers.}~~
Transformers have emerged as a promising alternative to convolutional neural networks (CNNs) as an architecture for computer vision tasks recently~\cite{dosovitskiy2020image,liu2021swin}. 
While CNNs have been extensively studied for robustness, the robustness of vision transformers are still under-explored.
Some works~\cite{bhojanapalli2021understanding,mahmood2021robustness} have shown that transformer architecture maybe more robust to adversarial examples, but it remains if this result holds for OOD robustness.
In the following, we compare the performance of CNNs and transformers on the tasks of image classification, object detection and 3D pose estimation on the \OURS benchmark.
Specifically, we replace the backbone the vision models for each task from ResNet-50 to Swin-T~\cite{liu2021swin}. 
Our experimental results are presented in~\cref{fig:cnn_vit}. Each experiment is performed five times and we report mean performance and standard deviation.
It can be observed that CNNs and vision transformers have a comparable performance across all tasks as the difference between their performances are within the margin of error. 
Particularly, we do not observe any enhanced robustness as OOD shifts in individual nuisance factors lead to a similar decrease in performance in both the transformer and the CNN architecture.
While we observe a slight performance gain on i.i.d. data in image classification (as reported in many other works), our results suggest that \textbf{Transformers do not have any enhanced OOD robustness compared to CNNs}.
Note our findings here contrast with previous work on this topic~\cite{bai2020vitsVScnns}, we argue that this is because our benchmark enables the study for individual nuisance factors on real world images, and the control over different individual nuisances give us opportunity to observe more errors in current vision models.


\noindent \textit{Model capacity.}~~~
For deployment in real applications, smaller models are preferred because they can yield better efficiency than regular models.
In the following, we compare image classification performance of MobileNetV3~\cite{howard2019searching_mbv3} in~\cref{tab:mbv3_vs_r50}.
Compared to ResNet-50, MobileNetv3 suffers a similar performance degradation under OOD shifts in the data.
However, data augmentations does not improve the robustness of MobileNetV3~\cite{howard2019searching_mbv3} as much as for ResNet-50, \eg, performance on context nuisances improved by $3.9\%$ for ResNet-50, but the improvement is only $0.9\%$ for MobileNetV3.  
This suggests that \textbf{OOD robustness is more difficult to achieve for efficient models with a limited capacity}.


\begin{table*}
\footnotesize
\centering
\caption{\label{tab:one_two_stage} Comparison between one-stage method and two-stage object detection methods. One-stage methods are more robust compared to two-stage methods.}
\setlength\tabcolsep{0.3em}
{\color{black}
\begin{tabular}{lccccccc}
\toprule
                                     & i.i.d & shape & pose & texture & context & weather & occlusion \\ 
\midrule
RetinaNet~\cite{lin2017focal}        & 43.9\% & 39.2\% & 33.4\% & 37.4\% & 27.8\% & 31.3\% &  16.8\%\\
+Style Transfer~\cite{geirhos2018}   & 44.3\% & 38.7\% & 32.1\% & 39.2\% & 31.2\% & 34.6\% &  17.3\% \\
\midrule
Faster-RCNN~\cite{ren2015faster}     & 40.6\% & 34.9\% & 30.4\% & 34.6\% & 27.0\% & 28.3\% &  10.0\% \\
+Style Transfer~\cite{geirhos2018}   & 41.9\% & 36.4\% & 31.6\% & 37.9\% & 31.1\% & 33.3\% &  14.0\% \\ 
\bottomrule
\end{tabular}
}
\end{table*}

\begin{table*}
\small
\centering
\caption{\label{tab:pose_nemo_r50}
Robustness of 3D pose estimation methods. 
We compare ``Res50-Specific'', which treats pose estimation as classification problem, and ``NeMo'', which represents the 3D object geometry explicitly.
We observe OOD shifts in shape and pose leads to more performance degradation. NeMo has a significantly enhanced performance to OOD shifts in object shape and pose.
}
\setlength\tabcolsep{0.3em}
{\color{black}
\begin{tabular}{lccccccc}
\toprule
                                     & i.i.d   & shape  & pose   & texture & context & weather & occlusion  \\ 
\midrule
Res50-Specific                     & 62.4\% & 43.5\% & 45.2\% & 51.4\% & 50.8\% & 49.5\%  &  41.6\% \\
+AugMix~\cite{hendrycks2019augmix} & 64.8\% & 44.1\% & 44.8\% & 56.7\% & 54.7\% & 55.6\%  &  42.1\% \\
\midrule
NeMo~\cite{wang2021nemo}           & 66.7\% & 51.7\% & 56.9\% & 52.6\% & 51.3\% & 49.8\%  &  50.8\% \\
+AugMix~\cite{hendrycks2019augmix} & 67.9\% & 53.1\% & 58.6\% & 57.8\% & 55.1\% & 56.7\%  &  52.2\% \\
\bottomrule
\end{tabular}
}
\end{table*}

\noindent \textit{One stage vs two stage for detection.}~~~
It is a common belief in object detection community that two-stage detectors are more accurate, while one-stage detectors are more efficient.
For object detection task, two popular types of architecture exist, namely one-stage and two stage models.
We tested two representative models from these architecture types, RetinaNet~\cite{lin2017focal}, a one-stage detector, and Faster-RCNN~\cite{ren2015faster}, which is a two-stage detector.
From our results in~\cref{tab:one_two_stage}, we observe that RetinaNet achieves a higher performance compared to Faster-RCNN on the \OURS benchmark.
However, when accounting for improved i.i.d performance, the OOD performance degradation are similar between two models.
These initial result suggests that \textbf{two-stage methods achieve a higher score than one-stage methods, but are not necessarily more robustness}. 

\noindent \textit{Models with explicit 3D object geometry.}~~~
Recently, Wang et al.~\cite{wang2021nemo} introduced NeMo, a neural network architecture for 3D pose estimation that explicitly models 3D geometery, and they demonstrated promising results on enhancing robustness to partial occlusion and unseen 3D poses.
In~\cref{tab:pose_nemo_r50}, we compare NeMo~\cite{wang2021nemo} model and a general Res50-Specific model on task of pose estimation on \OURS benchmark.
NeMo~\cite{wang2021nemo} shows a stronger robustness against geometric-based nuisances (shape and pose), while robustness on appearance-based nuisances is comparable.
This result suggests that, \textbf{neural networks with an explicit 3D object representation have a largely enhanced robustness to OOD shifts in geometry-based nuisances}. 
These results seem complementary to our experiments in the previous section, which demonstrate that strong data augmentation can help to improve the robustness of vision models to appearance-based nuisances, but not to geometry-based nuisances.

We further investigate, if robustness against all nuisance types can be improved by combining data augmentation with architectures that explicitly represent the 3D object geometry. Specifically, we train NeMo~\cite{wang2021nemo} with strong augmentations like AugMix~\cite{hendrycks2019augmix} and our results in~\cref{tab:pose_nemo_r50} show that this indeed largely enhances the robustness to OOD shifts in appearance-based nuisances, while retaining (and slightly improving) the robustness to geometry-based nuisances.
Result suggests that enhancements of robustness to geometry-based nuisances can be developed independently to those for appearance-based nuisances.

\subsection{OOD shifts in Multiple Nuisances}
\label{sec:combined_nuisance}
In our experiments, we observed that geometry-based nuisances have different effects compared to appearance-based nuisances.
In the following, we test the effect when OOD shifts happen in both of these nuisance types. 
Specifically, we introduce new dataset splits, which combine appearance-based nuisances, including texture, context, or weather, with the geometry-based nuisances shape and pose. From~\cref{tab:shape_pose_plus_x}, we observe \textbf{OOD shifts in multiple nuisances amplify each other}.
For example, for image classification, an OOD shift in only the 3D pose reduces the performance by $-11.4\%$ from $85.2\%$ to $73.8\%$, and an OOD shift in the context reduces the performance by $-6.6\%$.
However, when pose and context are combined the performance reduces by $-24.5\%$. 
We observe a similar amplification behaviour across all three tasks, suggesting that it is a general effect that is likely more difficult to address compared to single OOD shifts.


\begin{table}
\centering
\caption{\label{tab:shape_pose_plus_x} Robustness to OOD shifts in multiple nuisances. When combined, OOD shifts in appearance-based nuisances and geometric-based nuisances amplifies each other, leads to further decrease compared to effects in individual nuisances.}
{\color{black}
\begin{tabular}{lccccc}
\toprule
                & i.i.d             & texture   & context       & weather  & occlusion  \\ 
\midrule
Classification  & 83.9\%           & 67.6\%    & 65.1\%        & 71.5\%   & 58.5\%   \\
+ shape    & 68.7\%           & 45.7\%    & 44.0\%        & 48.3\%  & 36.7\%   \\
+ pose     & 76.1\%           & 47.1\%    & 43.5\%        & 51.4\%   & 40.1\%   \\
\midrule
Detection       & 40.6\%           & 34.6\%    & 27.0\%        & 28.3\%   & 10.0\%   \\
+ shape    & 34.9\%           & 10.6\%    & 8.7\%        & 9.4\%   & 3.7\%   \\
+ pose     & 30.4\%           & 10.0\%    & 7.6\%        & 8.9\%   & 5.4\%   \\
\midrule
Pose estimation & 62.4\%    & 51.4\%    & 50.8\%        & 49.5\%     & 41.6\%   \\
+ shape         & 43.5\%    & 33.1\%    & 31.0\%        & 29.8\%     & 26.3\%  \\
+ pose          & 45.2\%    & 30.2\%    & 29.7\%        & 28.1\%     & 24.7\%   \\
\bottomrule
\end{tabular}
}
\end{table} 

{\color{black}
\subsection{OOD Shifts in Partial Occlusion}
\label{sec:exp:occ}
Partial occlusion is a challenging novel OOD shift that we added to the OOD-CV dataset for this journal extension (see examples in Figure \ref{fig:data_demo}. Our comprehensive experiments demonstrate that occlusion is the nuisance that causes the largest drops in performance across all tasks image classification, object detection and 3D pose estimation (Table \ref{tab:individual_nuisance_on_different_tasks}).
Moreover, data augmentation like StyleTransfer, AugMix and PixMix only have marginal effects on the robustness to partial occlusion across vision tasks (Table \ref{tab:data_aug}) except for object detection.
Moreover, the architectural changes between CNNs and Transformers also only show a rather small effect on the robustness to partial occlusion.
A more promising approach to enhance robustness to partial occlusion can be observed in the context of 3D pose estimation in Table \ref{tab:pose_nemo_r50}, where the generative model NeMo significantly outperforms the ResNet50 baseline.
This confirms observations made in prior works \cite{kortylewski2020compositional, kortylewski2021compositional, wang2020robust, wang2021nemo}
that neural network architectures which replace the classic fully-connected prediction heads with a generative model of the neural feature activations, have a largely enhance occlusion robustness, because they can localize occluders and subsequently focus on the non-occluded parts of the object.
}

{\color{black}
\subsection{Consistent Classification and Pose estimation}
\label{sec:exp:3dclass}
In this section, we refer to Consistent Classification and Pose estimation (CCP) \cite{classNemoICCV} as the joint estimation of the 3D pose of the object and its class label. 
We evaluate CCP performances when several nuisances occur. Our results in Table \ref{tab:3daware_table} show the complexity of performing CCP w.r.t. OOD shifts in six nuisance factors. Current state-of-the-art never exceeds $50\%$ accuracy in any OOD scenarios when it performs up to an accuracy of almost $74\%$ in IID scenarios. Additionally, we observe consistently better performances of generative-based approaches compared to non-generative-based approaches. Over all nuisances, table \ref{tab:3daware_table} shows 2x up to 4x improvements of generative-based approach RCNet for Acc-${\frac{\pi}{6}}$ and Acc-${\frac{\pi}{18}}$ respectively, once again demonstrating the enhanced robustness of models with a 3D object representation. The conformity of these results with prior discoveries may be attributed to the inherent resilience of 3D-aware and generative-based approaches towards disturbances.}

\begin{table}
  \tabcolsep=0.11cm
  \caption{\color{black} Consistent Classification and Pose estimation (CCP) results on OOD-CV dataset. We observe a big performance degradation when performing complex tasks such as CCP in OOD scenarios. However, we see that methods using a 3D object representation as prior (e.g., RCNet) outperforms considerably non-generative approaches (e.g., Resnet50)\cite{classNemoICCV}.}
    \label{tab:3daware_table}
    \begin{subtable}{\linewidth}
    {\color{black}
    \begin{tabular}{lccccccc}  
    \toprule
        Acc-${\frac{\pi}{6}}\uparrow$ & i.i.d & shape & pose & texture & context & weather & occlusion  \\ 
    \midrule
    Resnet50& 73.9 & 15.7 & 12.6 & 22.3 & 15.5 & 23.4 & 24.9 \\
        
    
    RCNet & 85.8 & 52.9 & 21.5 & 55.4 & 50.2 & 55.3 & 59.2\\
    \end{tabular}
    
    \begin{tabular}{lccccccc}  
    \midrule\midrule
        Acc-${\frac{\pi}{18}}\uparrow$ & i.i.d & shape & pose & texture & context & weather & occlusion\\ 
    \midrule
    Resnet50& 45.6  & 5.7 & 5.7 & 4.3 & 6.1 & 5.4 & 12.4\\

     
    
    RCNet & 61.5 & 19.7 & 8.2 & 28.6 & 20.8 & 34.5 & 27.0 \\
    
    \bottomrule
    \end{tabular}
    }
    \end{subtable}
\end{table}
\color{black}{
\subsection{Robust 6D Pose Estimation}
Category-level 6D pose estimation involves joint 3D object detection and pose estimation. 6D pose estimation is evaluated by both the pose error and the average distance metric (ADD). We evaluate the 6D pose estimation performance under various nuisances and the results are reported in Table~\ref{tab:6pose_table}. We notice significant drop in performance for state-of-the-art category-level 6D pose estimation methods such as Faster R-CNN \cite{ren2015faster} and RTM3DExt \cite{li2020rtm3d}, under various OOD scenarios. Moreover, we find that C2F-NF \cite{ma2022robust} with a 3D neural feature representation demonstrates a comparable or stronger robustness for all nuisances. This is consistent with the findings in other tasks that generative methods with 3D geometry representations are more robust to the nuisances considered in our OOD-CV dataset.

\begin{table}
  \tabcolsep=0.11cm
  \caption{\color{black} Robust 6D pose estimation on OOD-CV dataset. When evaluated on OOD scenarios, we notice a clear degradation of performance for all state-of-the-art models. We also find that C2F-NF with a 3D neural feature representation demonstrates a comparable or stronger robustness for all nuisances.}
    \label{tab:6pose_table}
    {\color{black}
    \resizebox{\linewidth}{!}{%
    \begin{tabular}{lccccccc}   
    \toprule
    Acc-$\frac{\pi}{6} \uparrow$ & i.i.d. & context & occlusion & pose & shape & texture & weather \\ 
    \midrule
    FRCNN & 52.4 & 35.3 & 33.9 & 14.6 & \textbf{40.0} & 43.0 & 41.3 \\
    RTM3DExt & 46.1 & 34.5 & 34.9 & 11.7 & 35.2 & 40.7 & 39.5 \\
    C2F-NF & \textbf{58.7} & \textbf{37.2} & \textbf{45.4} & \textbf{15.9} & 39.4 & \textbf{46.3} & \textbf{46.5} \\
    \midrule\midrule
    Acc-$\frac{\pi}{18} \uparrow$ & i.i.d. & context & occlusion & pose & shape & texture & weather \\ 
    \midrule
    FRCNN & 20.5 & 10.8 & 10.3 & 3.5 & \textbf{11.8} & 16.3 & 18.7 \\
    RTM3DExt & 16.2 & 13.6 & 11.4 & 2.2 & 10.4 & 15.1 & 14.8 \\
    C2F-NF & \textbf{25.7} & \textbf{13.7} & \textbf{13.7} & \textbf{15.9} & 11.7 & \textbf{18.4} & \textbf{23.0} \\
    \midrule\midrule
    Med-ADD $\downarrow$ & i.i.d. & context & occlusion & pose & shape & texture & weather \\ 
    \midrule
    FRCNN & \textbf{0.64} & 1.82 & \textbf{1.26} & 2.26 & 0.65 & \textbf{1.38} & \textbf{1.02} \\
    RTM3DExt & 1.92 & 2.21 & 2.95 & 2.74 & 1.30 & 2.91 & 2.13 \\
    C2F-NF & 0.85 & \textbf{1.26} & 2.06 & \textbf{0.52} & \textbf{0.47} & 2.09 & 1.84 \\
    \bottomrule
    \end{tabular}
    }
    }

\end{table}

}

{\color{black}

\subsection{Summary and Discussion}
\label{sec:summary_disc}
Our results highlight a general and fundamental research problem that is inherent to current vision algorithms: \textbf{A lack of robustness to OOD shifts in the data for all state-of-the-art vision models across several important computer vision tasks.}
Going beyond prior works, our OOD-CV dataset enables us to study the effect of OOD shifts in individual nuisance factors in real images for several vision tasks.
%
%
One important observation of our experiments is that the nuisance factors have a different effect on different vision tasks, suggesting that each vision task might need a different solution for enhancing the OOD robustness.
From our experiments we can also clearly observe that the nuisance variations can be categorized into two sets: \textit{appearance-based nuisances} like texture, context, or weather, and \textit{geometry-based nuisances} such as shape or pose. 
We showed that strong data augmentation enhances the robustness against appearance-based nuisances, but has very little effect on geometric-based nuisances. 
On the other hand, neural network architectures with an explicit 3D object representation achieve an enhanced robustness against geometric-based nuisances.
While we observe that OOD robustness is largely an unsolved and severe problem for computer vision models, our results also suggest a way forward to address OOD robustness in the future. Particularly, that approaches to enhance the robustness may need to be specifically designed for each vision tasks, as different vision tasks focus on different visual cues.
Moreover, we observed a promising way forward to a largely enhanced OOD robustness is to develop neural network architectures that represent the 3D object geometry explicitly and are trained with strong data augmentation to address OOD shifts in both geometry-based and appearance-based nuisances combined.


}
\section{Conclusion}
\label{sec:summary_disc}

{\color{black}
Our study makes several major contributions: 

1) We raise attention for the fundamentally important problem of out-of-distribution robustness, and the pressing issues it implies for autonomous agents that shall interact within a real-world scenarios. 

2) We introduce the first benchmark for out-of-distribution robustness with real images and detailed annotation of nuisance variables for various important vision tasks.

3) Despite being largely acknowledged, progress in OOD robustness is limited as highlighted in our study.
Based on our results, a promising way forward to resolve this fundamental problem is to design neural network architectures that have explicit 3D representations of objects to generalize under geometry-based OOD shifts, paired with advanced data augmentation to enhance appearance-based OOD robustness.
}

\ifCLASSOPTIONcompsoc
  \section*{Acknowledgments}
\else
  \section*{Acknowledgment}
\fi
AK acknowledges support via his Emmy Noether Research Group funded by the German Science Foundation (DFG) under Grant No. 468670075.
AY acknowledges grants ONR N00014-20-1-2206 and ONR N00014-21-1-2812.

{
\bibliographystyle{ieee_fullname}
\bibliography{eccv_egbib}
}

\clearpage

\appendices

\section{}
\textbf{Images filtered from the original PASCAL3D+ dataset}
This section shows example images that we filtered out from the original PASCAL3D+ dataset~\cite{everingham2015pascal} in order to make the \OURS test set really OOD.
The images are removed because they are too similar to the images in the \OURS test set.

In our anonymous repository, we provide all the images that we removed from the original PASCAL3D+ dataset.

\begin{figure}
\includegraphics[width=\linewidth]{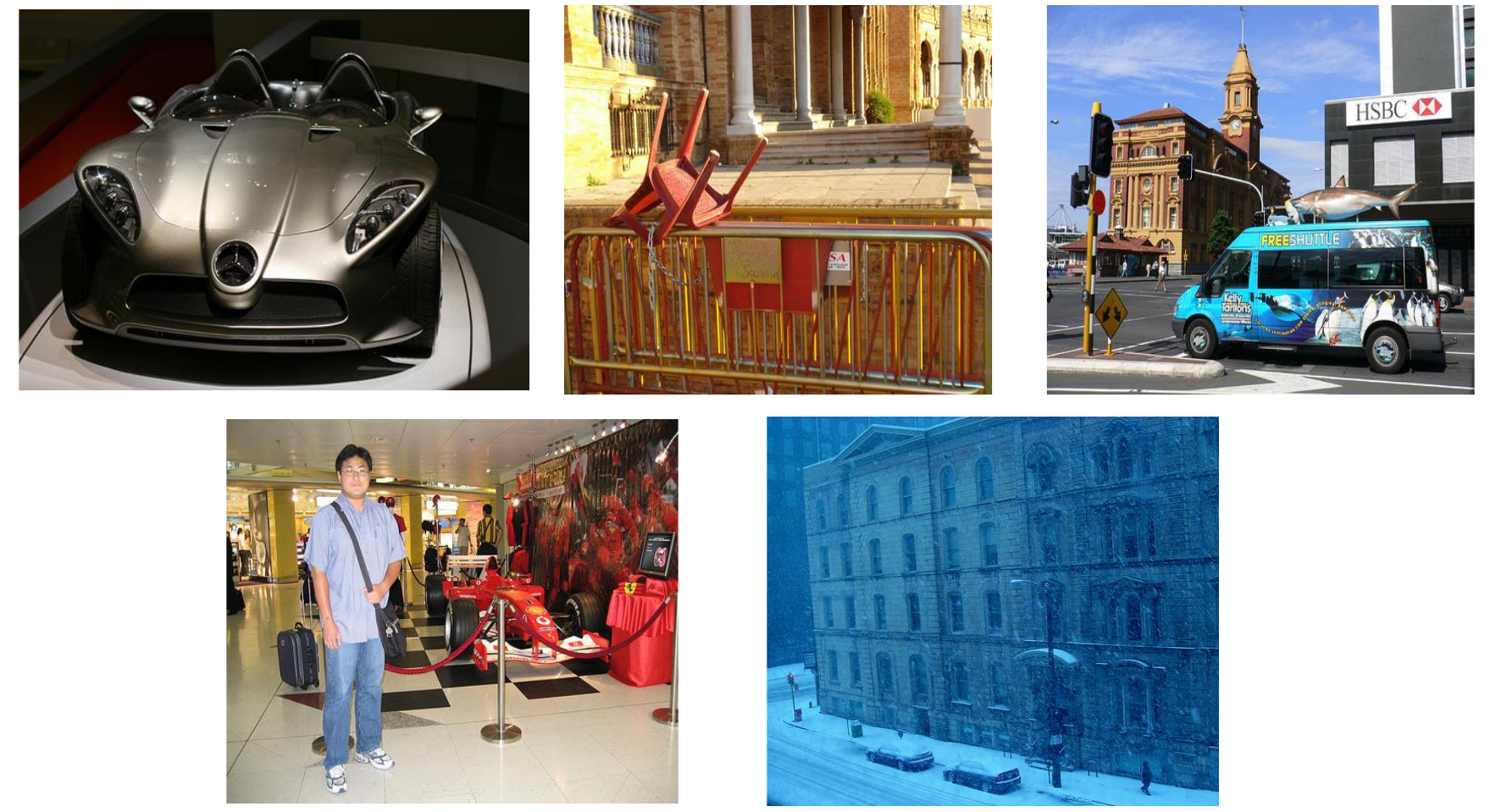}
\caption{\label{fig:filter_pascal} Example images that are filtered out from the original PASCAL3D+ dataset. These images has nuisances that are similar to the ones we collected in the \OURS dataset, so they are removed from the training set. }
\end{figure}

\begin{figure}
\includegraphics[width=\linewidth]{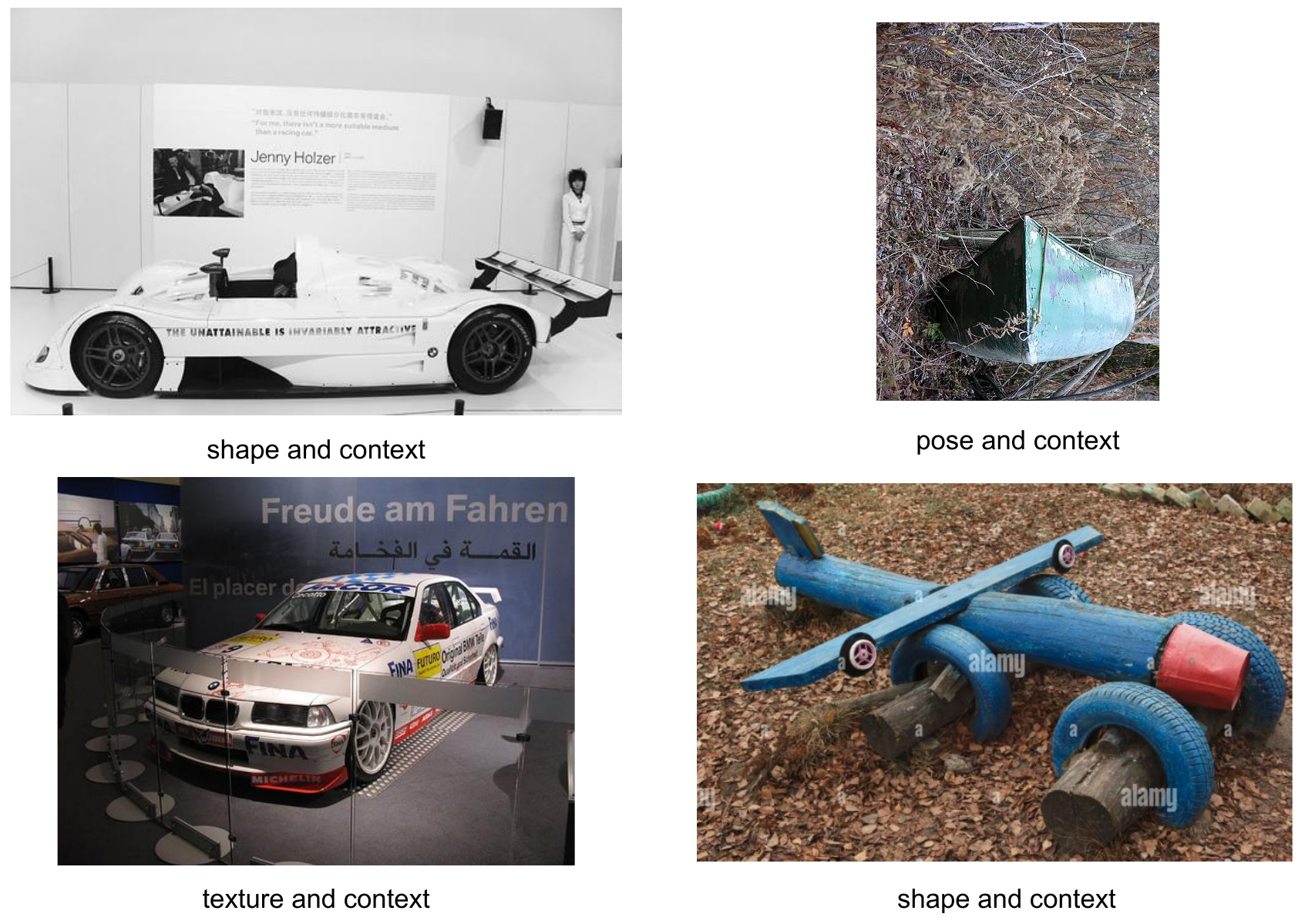}
\caption{\label{fig:multiple_nuisance} Example images with multiple nuisance. From our internet search, we also collected many images with multiple nuisance factors, these images are later removed to ensure that we are testing with only one controllable nuisances.}
\end{figure}

\textbf{Example images with multiple nuisances}
We also removed the images that have multiple nuisances from our internet search, we give examples of multiple nuisances in~\cref{fig:multiple_nuisance}.

\textbf{The user interface of our annotation tools}
Here we also provide the user interface of our used annotation tools for bounding boxes annotation and 3D pose annotations.
The annotation tools are taken and slightly modified from a GitHub project~\footnote{https://github.com/jsbroks/coco-annotator} and the original PASCAL3D+ dataset~\footnote{https://cvgl.stanford.edu/projects/pascal3d.html}. Identifying informations have been removed from the screenshots.

\begin{figure}[h]
\includegraphics[width=\linewidth]{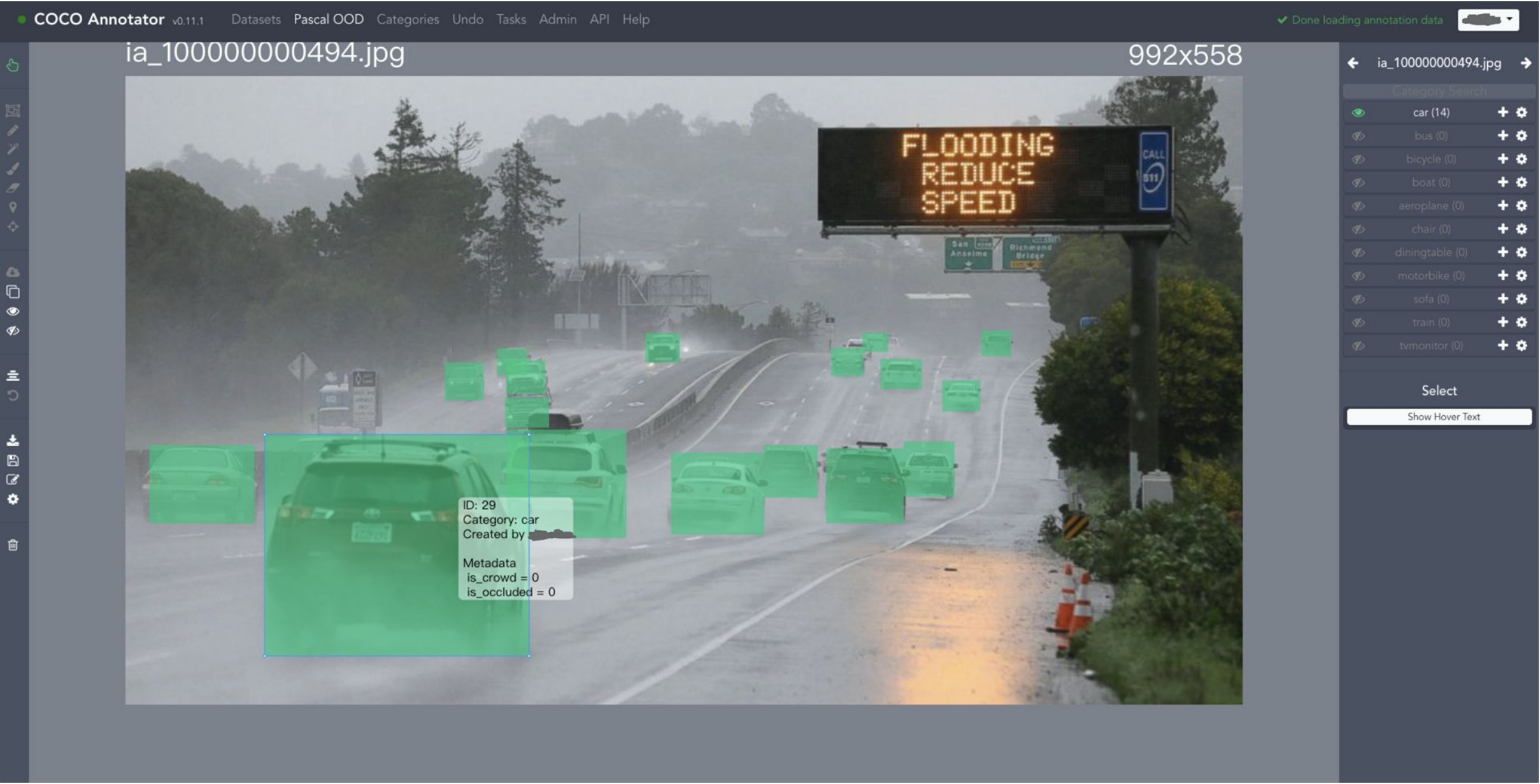}
\caption{\label{fig:det_ui} The user interface of the detection annotation tool.}
\end{figure}

\begin{figure}[h]
\includegraphics[width=\linewidth]{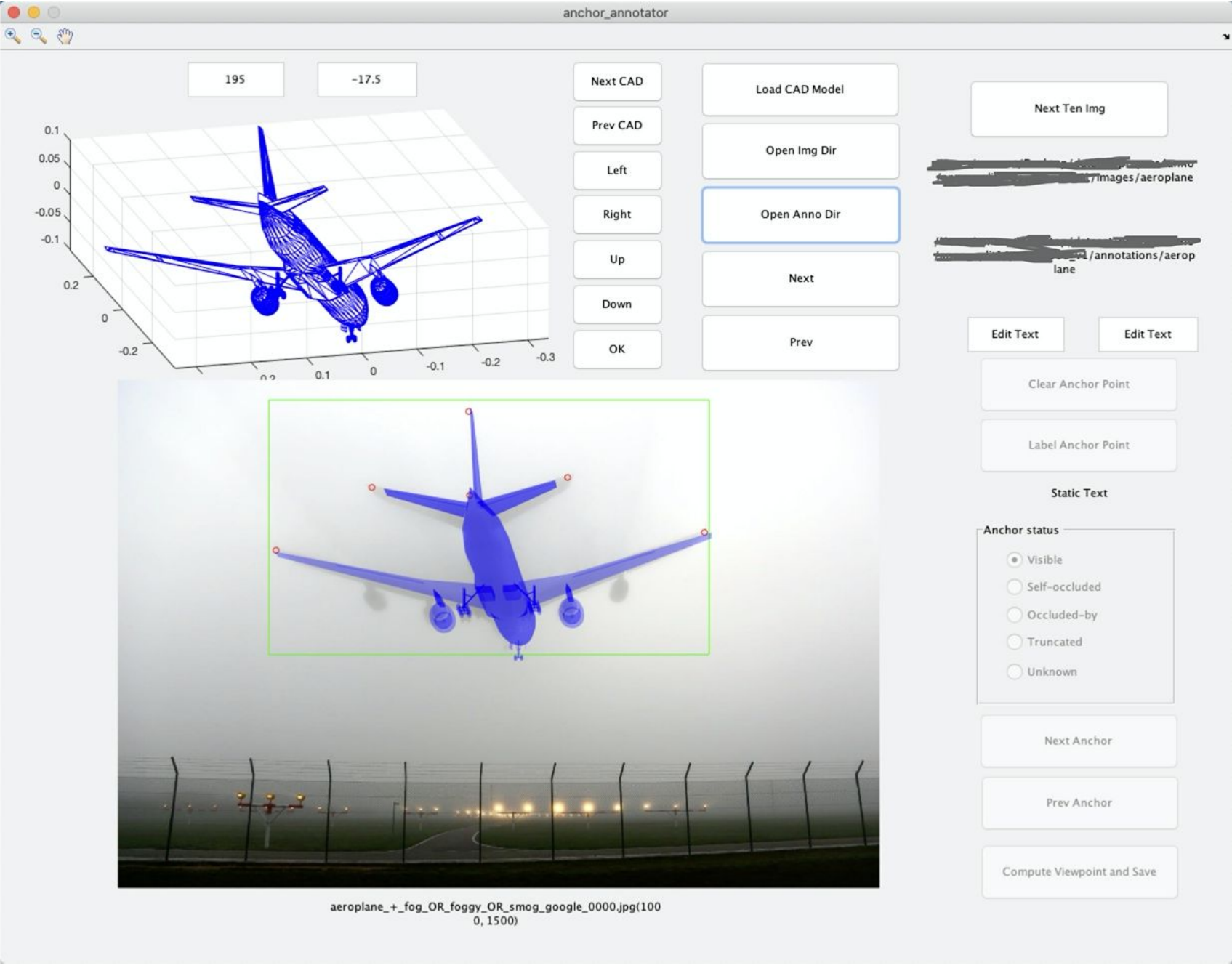}
\caption{\label{fig:det_ui} The user interface of the 3D pose annotation tool.}
\end{figure}

\section{}
This section provides information on a paper that is currently unpublished but has been referenced in our main paper\cite{classNemoICCV}. This unpublished paper contains relevant and supplementary details that support the findings and conclusions presented in our research.


\section*{Supplementary material}

# Robust Object Classification via Render-and-Compare with 3D-aware Deep Networks

Artur Jesslen[1*]  Guofeng Zhang[2*]  Angtian Wang[3]  Alan Yuille[3†]  Adam Kortylewski[1,4†]

[1]University of Freiburg  [2]University of California, Los Angeles
[3]Johns Hopkins University  [4]Max-Planck-Institute for Informatics

March 31, 2023

## Abstract

In real-world applications, it is essential for deep networks to classify objects robustly in out-of-distribution (OOD) scenarios, i.e. when the test data does not come from the same distribution as the training data. Inspired by the robustness of render-and-compare approaches at 3D pose estimation, we propose a novel network architecture (termed RCNet) that follows a render-and-compare approach and achieves largely enhanced robustness at object classification. RCNet represents an object category as a 3D cuboid mesh composed of feature vectors at each mesh vertex. During inference, the model searches for the category mesh and corresponding 3D pose that best reconstructs a target feature map. Importantly, the neural textures on each mesh are trained in a discriminative manner to enable classification. To achieve an efficient inference the model combines the 3D-aware head with a standard (non-robust) fully-connected head while retaining robustness. Our experiments show that RCNet is exceptionally robust on a range of real-world and synthetic distribution shifts while performing on par with state-of-the-art architectures on in-distribution data. Moreover, the predicted 3D pose of RCNet is competitive with baseline models that were explicitly designed for robust pose estimation.

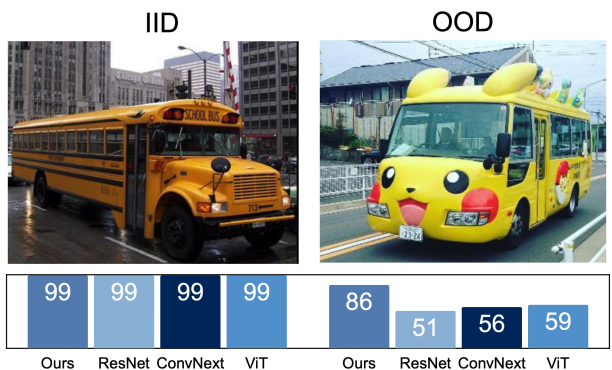

Figure 1: Current state-of-the-art architectures for object classification work well when the training and test data are drawn from a similar distribution (left), but their performance decreases significantly in out-of-distribution scenarios (right). In contrast, our proposed RCNet architecture achieves unprecedented robustness in OOD scenarios while performing on par with baseline models on IID data (results show object classification accuracy obtained on the OOD-CV dataset [40]).

---

*Joint first authors
†Joint senior authors

# 1 Introduction

Deep neural networks, achieve very high performance at image classification [7, 12, 20, 22, 30], under the implicit assumption that the training and test data are drawn independently and identically distributed (IID) from the same distribution. Unfortunately, this is not a realistic assumption in the real world. For example, having learned to classify buses from a large labeled dataset, a model will still encounter unusual buses in the wild that were not observed during training. As a practical consequence, the performance of today's best deep networks decreases significantly when evaluated in out-of-distribution (OOD) scenarios (Figure 1). Overcoming this lack of robustness is essential for deploying deep networks in safety-critical applications, and has been identified as a core open problem of deep learning, for example by Yoshua Bengio, Geoffrey Hinton, and Yann LeCun [2]. However, the problem largely remains unsolved.

In contrast to computer vision, human vision is highly robust and generalizes well under occlusion or environmental changes. Cognitive studies suggest that the robustness of human visual perception arises from an analysis-by-synthesis process [27, 39]. Current AI systems implement analysis-by-synthesis through a render-and-compare process, where a generative forward model (typically a graphics pipeline) is used together with an explicit 3D representation of the object (e.g., a CAD model) to generate images of an object class. During inference, a given test image is analyzed by searching for the model parameters that best reconstruct the image. Render-and-compare has found wide applicability in 3D and 6D pose estimation [4, 5, 15, 16, 21], where it was also shown to be exceptionally robust [24, 36]. However, for object classification, render-and-compare approaches are rare and understudied.

In this work, we introduce RCNet, a deep network architecture that implements a render-and-compare approach for object classification and achieves unprecedented robustness in out-of-distribution scenarios. Following prior works on render-and-compare that were designed for 3D pose estimation [16, 36], RCNet represents an object category as a cuboid mesh and learns a generative model of the neural feature activations at each mesh vertex. To classify images, RCNet first extracts a feature map of the target image and subsequently searches the category mesh and corresponding 3D pose that best reconstructs the target feature map. Intuitively, robustness in RCNet emerges, because it has a causal generative understanding of the 3D structure of the object, and hence can better generalize, e.g. to unseen views of an object or when parts are occluded.

We make several contributions to enable this render-and-compare approach to object classification. First, we train the neural textures of object categories using contrastive learning to be distinct between different classes, while also being invariant to instance-specific details within an object category. Second, to avoid computing the render-and-compare optimization for all category meshes, the model uses a fully-connected head (FC-head) that makes an initial (non-robust) prediction of the object class and pose. We show how the 3D-aware head can verify the output of the (non-robust) FC-head to reach a sweet spot at which the fast-to-compute FC-head is used when test data is easy to classify while falling back to the more expensive but reliable 3D-aware head in difficult OOD cases.

We evaluate RCNet under real-world OOD shifts on the OOD-CV dataset [40], and synthetic OOD shifts on the corrupted-PASCAL3D+ [13] and occluded-PASCAL3D+ [36]. Our experiments show that RCNet is exceptionally more robust compared to other state-of-the-art architectures (both CNNs and Transformers) at object classification while performing on par with in-distribution data. Moreover, we show that the predicted 3D pose of RCNet is competitive with baseline models that were explicitly designed for robust 3D pose estimation. Our contributions are as follows:

1. We introduce RCNet, a 3D-aware neural network architecture that follows a render-and-compare approach and leverages discriminatively trained neural textures to perform object classification.

2. We integrate the 3D-aware head with a standard (non-robust) fully-connected head to considerably speed up the render-and-compare inference while retaining classification accuracy and robustness.

3. We show that RCNet is exceptionally robust at object classification and performs competitively with SOTA methods at 3D pose estimation.

## 2 Related Work

**Robust Image Classification.** Image Classification is a significant task in computer vision. Multiple influential architectures including ResNet [12], Transformer [35], and recent Swin-Transformer [22] have been designed for this task. However, these models are not robust enough to handle partially occluded images or out-of-distribution data. Efforts that have been made to close the gap can be mainly categorized into two types: data augmentation and architectural design. Data augmentation includes using a learned augmentation policy [6], and data mixture [14]. Architectural changes propose robust pipelines. For instance, [18] proposes an analysis-by-synthesis approach for a generative model to handle occlusions. In addition, challenging benchmarks that use common synthetic corruptions [13] and real out-of-distribution images [40] showed that the performance of standard models drops by a large margin in such scenarios. We show that our RCNet model is exceptionally robust in these challenging OOD scenarios.

**Feature-level render-and-compare.** Render-and-compare methods optimize the predicted pose by reducing the reconstruction error between 3D-objects projected feature representations and the extracted feature representations. It can be seen as an approximate analysis-by-synthesis [11] approach, which has been proven to be much more robust against out-of-distribution data at 3D pose estimation [17, 36] compared to classical discriminative methods [26, 34, 41].

Inspired by these results, we introduce the first render-and-compare approach to object classification, that demonstrates exceptional robustness on a wide range of distribution shifts while performing on par with the most competitive image classification architectures.

**Combining feed-forward networks with generative models.** As the optimization process in render-and-compare methods is computationally expensive, some works combined feed-forward neural networks to speed up the inference process, particularly in the context of 3D face reconstruction [28, 31], where generative models are widely applied [9]. However, to the best of our knowledge, no prior work has studied the integration of deep networks and generative models to retain a fast but at the same time robust prediction output. We take inspiration from these approaches and combine the fast (but not robust) inference of a feed-forward fully-connected predictor with the robustness of our generative approach to obtain a fast and robust object classification within a render-and-compare approach.

## 3 RCNet: A 3D-aware Deep Network with Render-and-Compare

We first review the concept of feature-level rendering (Section 3.1). Subsequently, we describe in Section 3.2 how object classification using render-and-compare is implemented within the architecture of RCNet, and the RCNet architecture can be further extended by combining the proposed 3D-aware classification head with a (non-robust) fully-connected (FC) head to enhance inference time. We introduce a principled way of formulating the render-and-compare approach to object classification within a probabilistic framework in Section 3.3 and describe the training in detail in Section 3.4.

### 3.1 Feature-level Rendering with Neural Textures

Following the approach of deferred neural rendering [32] and applications of it in pose estimation [16, 36], we represent an object category $y$ as a mesh $M_y = \{v_n \in \mathbb{R}^3\}_{n=1}^N$ and a *neural texture* $T_y$ on the surface of the mesh $M_y$. The neural texture is stored as a matrix of feature vectors for each vertex on the mesh $T_y \in \mathbb{R}^{N \times c}$, with $c$ being the number of channels in the feature vector. Each mesh is a cuboid with a fixed number of vertices and a fixed scale. Hence every object category is represented as a tuple $O_y = \{M_y, T_y\}$ of mesh and corresponding neural texture. We can render the object mesh $O_y$ into a feature map using standard computer graphics rasterization:

$$F_y(\alpha) = \Re(O_y, \alpha), \qquad (1)$$

where $\alpha$ is the 3D pose of the mesh in the camera view. Ultimately, we aim to perform classification by comparing this rendered feature map to a target feature map $F$ of an input image as described in the following section.

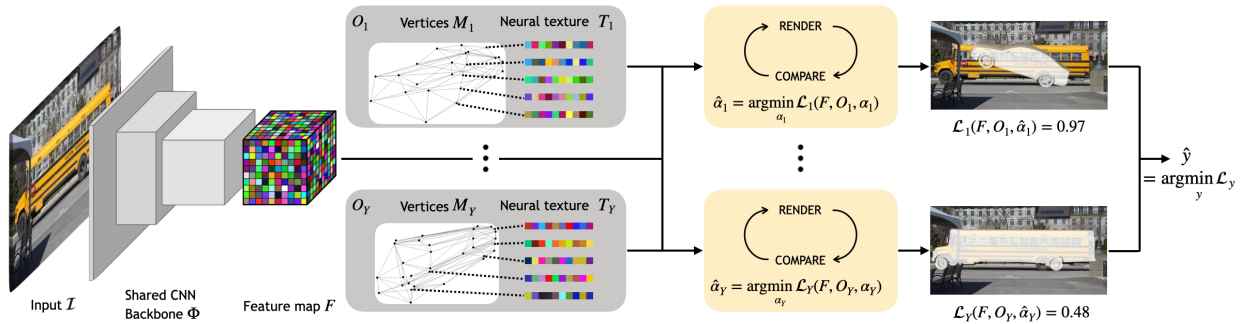

Figure 2: An overview of our RCNet pipeline. For each image, the network $\Phi$ extracts a feature map $F$. Meanwhile, we render feature maps $F'_y$ using our class-specific trained Neural Mesh Models $O_y$ and optimize the pose $\alpha_y$ using a feature-level render-and-compare approach. Lastly, we compare the different losses $\mathcal{L}_y(F, O, \alpha)$ to infer the object category $\hat{y}$. We visualize the pose prediction for two different object categories. Note that for illustration purposes, we visualize a projection of a CAD model but our RCNet is using cuboid meshes with much lower detailed geometry and we omit the background model $\mathcal{B}$ and the subscript $\mathcal{L}_{\text{Rec}}$ in the loss notation.

## 3.2 RCNet Architecture

In this section, we give a conceptual overview of the model architecture and inference process. Figure 2 illustrates the architecture and inference process of RCNet. An input image $\mathcal{I}$ is first processed by a shared deep network backbone $\Phi$ into a feature map $\Phi(I) = F \in \mathbb{R}^{H \times W \times D}$. The shared backbone is an important difference from other approaches that use feature-level rendering for 3D reconstruction [17, 36], as it enables our architecture to be trained discriminatively across classes. Based on the feature map $F$, the model minimizes the reconstruction loss (Eq. 3) through gradient-based optimization using a render-and-compare process to obtain the optimal pose $\hat{\alpha}_y$ for each object category $O_y$, and to determine the category $\hat{y}$ that achieves the lowest reconstruction error (hence minimizing Eq. 4). The key to achieving an accurate object classification using the RCNet architecture lies in training the feature extractor $\Phi$ for the optimal neural textures $\{T_y\}$ such that the reconstruction loss $\mathcal{L}_{\text{Rec}}$ indicates the correct class.

**Efficient Inference via Proposal Verification** A challenge in our render-and-compare approach to object classification is that a naïve implementation requires running the optimization for every object class, which is computationally expensive. We address this challenge by combining the 3D-aware classification head in RCNet with the output of a FC-head as illustrated in Figure 2. The FC-head predicts the corresponding probability for the object class $p(y|\mathcal{I}; \mathbf{w})$ and object pose $p(\alpha|\mathcal{I}; \mathbf{w})$ respectively for a given input image $I$, with $\mathbf{w}$ being the learned weights of the model. We propose a three-step process to integrate the 3D-aware head with the FC-head to achieve an efficient and robust prediction in OOD scenarios.

*(1) Handle simple cases using CNN predictions.* We take advantage of the ability of deep neural networks to approximately estimate the confidence of their prediction [10, 19] and retain the class prediction $\hat{y}_{\text{cnn}}$ if the confidence exceeds a threshold $p(y_{\text{cnn}}|I; \mathbf{w}) >= \tau_1$ with $\tau_1 = 0.95$ being experimentally defined. Subsequently, we fine-tune the initial pose prediction with the render-and-compare optimization to obtain a final result.

*(2) Verify feed-forward proposals with generative models in difficult cases.* In cases where the FC-head output is unreliable (i.e. $p(y_{\text{cnn}}|I; \mathbf{w}) < \tau_1$), its output is verified by the 3D-aware head in a "top-down" manner. Notably, such bottom-up and top-down processing was also advocated in classical prior works for object segmentation [3, 33] or face reconstruction [29]. In particular, we use the predicted top-3 classes and respective top-3 poses of the FC-head to compute the respective reconstruction losses using the 3D-aware head. We start a complete render-and-compare optimization process for each class

candidate, from the pose candidate that achieves the lowest reconstruction loss.

*(3) Resort to full render-and-compare optimization when uncertain.* The final reconstruction loss of the 3D-aware head can indicate if the render-and-compare optimization has converged to a good solution (Section 4). For those test images, where the reconstruction loss is above a threshold ($\tau_2 = 0.8$), we run the full inverse rendering process from several randomly initialized starting points and keep the best solution as described in Section 4.1.

This principled integration of the fast but not robust FC-head with our proposed 3D-aware head, significantly reduces the overall computation time compared to a naïve implementation of RCNet while retaining robustness. We note that this process is optional and only applies if a reduced computation time is desired.

### 3.3 Classification via Render-and-Compare

We formulate RCNet as a probabilistic generative model of neural feature activations. Following related work in 3D face reconstruction [28] we define the likelihood model as:

$$p(F|y) = p(F|O_y, \alpha_y, B) = \prod_{i \in \mathcal{FG}} p(f_i|O_y, \alpha) \prod_{i' \in \mathcal{BG}} p(f'_i|B), \quad (2)$$

where the foreground $\mathcal{FG}$ is defined as the area of the projected mesh and $\mathcal{BG}$ is the background context, and $B$ are the parameters for the background model.

Note that conditional independence makes it simple to "robustify" the likelihood with an outlier model [8, 18]. The foreground $p(f_i|O_y, \alpha)$ and background likelihoods $p(f_{i'}|B)$ are modeled as Gaussian distributions. Assuming unit variance, the negative log-likelihood reduces to [36]:

$$\mathcal{L}_{\text{Rec}}(F, O_y, \alpha_y, B) = -\log p(F|y)$$
$$= \sum_{i \in \mathcal{FG}} \|f_i - t_{y,n}\|^2 + \sum_{i' \in \mathcal{BG}} \|f'_i - B\|^2 + const. \quad (3)$$

Here, $t_{y,n}$ is the corresponding vertex $n$ on the object mesh that projects onto the pixel $i$ on the image plane, $\alpha$ are the camera extrinsic parameters, and $B \in \mathbb{R}^d$ are the parameters for the background model. We refer to this term as the *reconstruction loss*, since it measures the error between the target feature map and the rendered neural texture and background.

**Object Classification as Maximum Likelihood Estimation.** In contrast to prior work, which simply performed 3D reconstruction of a *known* object class within such a probabilistic framework [28, 36], our model needs to compare the reconstruction quality *across classes* to perform classification. This requires us to maximize the likelihood (i.e. minimize the reconstruction error) over the category label $y$ and the mesh pose $\alpha_y$ jointly, to find the mesh that best explains the target feature map $F$:

$$\{\hat{y}, \hat{\alpha}_{\hat{y}}\} = \arg\min_{y, \alpha_y} \mathcal{L}_{\text{Rec}}(F, O_y, \alpha_y, B), \quad (4)$$

where $\{\hat{y}, \hat{\alpha}_{\hat{y}}\}$ are the object class and corresponding object pose with minimal reconstruction error. This maximum likelihood inference process is implemented within the architecture of RCNet as described in Section 3.2.

### 3.4 Training RCNet

To train the feature extractor $\Phi$, the neural texture $\{T_y\}$ and the background model $B$ jointly, we utilize the EM-type learning strategy as originally introduced for keypoint detection in CoKe [1]. Specifically, the feature extractor is trained using stochastic gradient descent while the parameters of the generative model $\{T_y\}$ and $B$ are trained using momentum update after every gradient step in the feature extractor. Which was found to stabilize training convergence [1].

We use three training losses in total. First, the model parameters need to optimize the reconstruction loss $\mathcal{L}_{\text{Rec}}$ (Eq.3). To achieve this, we use the ground-truth 3D pose annotation to obtain correspondence between the mesh vertices and corresponding pixel locations in the respective feature maps as well as masks for the background $\mathcal{BG}$ regions in each training image.

We further use a contrastive loss [1] to maximize the distance between features far from each other on the 3D mesh, as well as between features on the object and the background. This contrastive property of features was found [36] to reduce the local optima in the feature-level reconstruction loss (Eq. 3), and hence enhances convergence in the render-and-compare process. To compute the loss, we first randomly sample a set of features from the

visible part of the object in the feature map $f_{u,v}, (u,v) \in \mathcal{FG}$. For each sampled feature $f_{u,v}$, we sample a set of negative training examples $(u', v') \in \mathcal{N} \subset \mathcal{FG}$ that are a certain distance away $\|(u', v') - (u, v)\|^2 > \tau$, where $\tau$ is a threshold for controlling the spatial discriminability. Together, with the features from the background region $\mathcal{BG}$ we compute a contrastive loss that maximizes the feature distance [36]:

$$\mathcal{L}_{con}(F, M_y, \alpha, B) = -\sum_{(u,v)} \sum_{(u',v')} \|f_{u,v} - f_{u',v'}\|^2 \\ - \sum_{i \in \mathcal{FG}} \sum_{j \in \mathcal{BG}} \|f_i - f_j\|^2, \quad (5)$$

which encourages the features at different positions on the object to be distinct from each other (first term) while also making the features on the object distinct from the features in the background (second term).

We further introduce a critical additional loss encouraging the neural textures among different meshes to be distinct from each other:

$$\mathcal{L}_{\text{class}}(\{T_y\}) = -\sum_{y=1}^{Y} \sum_{\bar{y}=1\setminus\{y\}}^{Y} \|\mu(y) - \mu(\bar{y})\|^2, \quad (6)$$

where $\mu(y) = \sum_{k=1}^{N} T_{k,y}/N$ is the mean of the neural texture $T_y$ of class $y$. Our full model is trained by optimizing the joint loss:

$$\mathcal{L}_{joint} = \mathcal{L}_{Rec} + \mathcal{L}_{con} + \mathcal{L}_{\text{class}}, \quad (7)$$

where the individual losses are weighted to be approximately in the same range.

**Training the FC-head.** After the training of the 3D-aware head and backbone feature extractor have converged, we train a fully-connected head on the output of the feature extractor using the class and pose label in the training data.

## 4 Experiments

In this section, we first describe the experimental setup in Section 4.1. Subsequently, we study the performance of RCNet in IID and OOD scenarios for different tasks in Sections 4.2-4.4. Finally, we perform an ablation for our 3D-aware head and our Speed-up approach in Sections 4.5-4.6.

### 4.1 Setup

**Datasets**. We evaluate RCNet on four datasets: PASCAL3D+ (P3D+) [38], occluded-PASCAL3D+ [37], corrupted-PASCAL3D+ [13, 25], and Out-of-Distribution-CV (OOD-CV) [40]. PASCAL3D+ includes 12 object categories, and each object is annotated with 3D pose, object centroid, and object distance. The dataset provides a 3D mesh for each category. We split the dataset into a training set of 11045 images and a validation set with 10812 images, referred to as L0. Building on the PASCAL3D+ dataset, the occluded-PASCAL3D+ dataset is a benchmark that evaluates robustness under different levels of occlusion. It simulates realistic occlusion by superimposing occluders on top of the objects with three different levels: L1: 20%-40%, L2: 40%-60%, and L3:60%-80%. corrupted-PASCAL3D+ corresponds to PASCAL3D+ on which we apply 12 types of corruptions [13, 25] to each image of the original P3D+ test dataset. We choose a severity level of 4 out of 5 for each applied corruption. Note that for fairness in the comparison, we didn't include any noise corruptions since the latest baselines (e.g., ConvNext, ViT) include some data pre-processing that reduces the noise level. The OOD-CV dataset is a benchmark dataset that includes OOD examples of 10 object categories varying in terms of 5 nuisance factors: pose, shape, context, texture, and weather.

**Implementation details.** RCNet consists of a category-specific neural mesh and a shared feature extractor backbone. Each mesh contains approx. 1100 vertices that are distributed uniformly on the cuboid. The shared feature extractor $\Phi$ is a ResNet50 model with two upsampling layers. The size of the feature map $F$ is $\frac{1}{8}$ of the input size. All images are cropped or padded to $640 \times 800$. The feature extractor and neural textures of all object categories are trained collectively, taking $\sim 20$ hours on 8 RTX 2080Ti.

RCNet inference follows Section 3.2. We extract a feature map $F$ from an input image using our shared feature extractor, then apply render-and-compare to render each neural mesh into a feature map $F'_y$. For initializing the pose estimation, we follow [36] and sample 144 poses (12 azimuth angles, 4 elevation angles, 3 in-plane rotations) and choose the pose with the lowest reconstruction error as initialization. We minimize the reconstruction loss of

| Dataset   | P3D+ | occluded-P3D+ | | | | OOD-CV | | | | | |
|-----------|------|------|------|------|------|---------|------|-------|---------|---------|------|
| Nuisance  | L0   | L1   | L2   | L3   | Mean | Context | Pose | Shape | Texture | Weather | Mean |
| Resnet50  | 99.3 | 93.8 | 77.8 | 45.2 | 79.6 | 45.1 | 61.2 | 55.2 | 48.3 | 47.3 | 51.4 |
| Swin-T    | **99.4** | 93.6 | 77.5 | 46.2 | 79.7 | 63.0 | 71.4 | 65.9 | 61.4 | 59.6 | 64.2 |
| Convnext  | **99.4** | 95.3 | 81.3 | 50.9 | 81.8 | 53.6 | 61.2 | 60.8 | 57.2 | 47.1 | 56.0 |
| ViT-b-16  | 99.3 | 94.7 | 80.3 | 49.4 | 80.9 | 57.8 | 67.3 | 61.0 | 54.7 | 54.5 | 59.0 |
| Ours      | 99.1 | 96.1 | 86.8 | 59.1 | 85.3 | **85.2** | **88.2** | **84.6** | **90.3** | 82.4 | **86.0** |
| Ours++    | **99.4** | **96.8** | **87.2** | **59.2** | **85.7** | 85.1 | 88.1 | 84.1 | 88.5 | **82.7** | 85.4 |

Table 1: Classification accuracy results on PASCAL3D+, occluded-PASCAL3D+ and OOD-CV datasets. L0 corresponds to unoccluded images from Pascal3D+, and occlusion levels L1-L3 are from occluded-PASCAL3D+ dataset with occlusion ratios stated in 4.1. RCNet performs similarly in IID scenarios, while steadily outperforming all baselines in OOD scenarios.

| Dataset   | P3D+ | corrupted-P3D+ | | | | | | | | | | | | |
|-----------|------|---------|---------|---------|---------|------|-------|------|------------|----------|------------------|---------|------|------|
| Nuisance  | L0   | defocus blur | glass blur | motion blur | zoom blur | snow | frost | fog | brightness | contrast | elastic transform. | pixelate | jpeg | mean |
| Resnet50  | 99.3 | 67.6 | 41.4 | 73.5 | 87.5 | 84.4 | 84.3 | 93.9 | 98.0 | 90.0 | 46.4 | 82.1 | 95.5 | 78.7 |
| Swin-T    | **99.4** | 60.7 | 37.1 | 70.9 | 81.3 | 88.5 | 91.6 | 95.4 | 97.9 | 92.1 | 56.3 | 79.2 | 95.3 | 78.9 |
| Convnext  | **99.4** | 70.1 | 58.7 | 76.5 | **90.0** | **92.3** | 92.9 | **98.5** | **99.2** | **98.4** | 67.6 | 84.2 | **98.7** | 85.6 |
| ViT-b-16  | 99.3 | 64.5 | **78.1** | 80.3 | 88.2 | 91.2 | **94.1** | 90.5 | 98.7 | 85.1 | 84.8 | **96.9** | **98.7** | 87.6 |
| Ours      | 99.1 | **90.1** | 66.9 | 86.8 | 84.9 | 81.3 | 88.1 | 98.2 | 97.9 | 96.8 | **96.7** | 96.9 | 98.1 | 90.2 |
| Ours++    | **99.4** | 89.9 | 66.4 | **87.3** | 87.2 | 83.3 | 89.8 | 98.4 | 98.0 | 96.9 | 96.5 | 96.7 | 98.4 | **90.4** |

Table 2: Classification accuracy results on corrupted-PASCAL3D+ under 12 different types of common corruptions. RCNet outperforms the baseline network that is was built upon (ResNet50) by a wide margin, and also outperforms other state-of-the-art architectures (even though RCNet is trained from plain images only, without any data augmentation).

| Dataset | P3D+ | occluded-P3D+ | corrupted-P3D+ | OOD-CV |
|---|---|---|---|---|
| Resnet50 | 39.0 | 15.8 | 15.8 | 18.0 |
| Swin-T | 46.2 | 16.6 | 15.6 | 19.8 |
| Convnext | 38.9 | 14.1 | 24.1 | 19.9 |
| ViT-b-16 | 38.0 | 15.0 | 21.3 | 21.5 |
| NeMo | 62.9 | **30.1** | 43.4 | 21.9 |
| Ours | 61.6 | 27.2 | 43.8 | **25.5** |
| Ours++ | **65.1** | 28.8 | **43.9** | 24.8 |

Table 3: Pose Estimation results for different datasets. A prediction is considered correct if the angular error is lower than $\frac{\pi}{18}$. Higher is better. Even though RCNet was not designed for robust 3D pose estimation, it performs similarly to the current state-of-the-art.

| Dataset | P3D+ | occluded-P3D+ | corrupted-P3D+ | OOD-CV |
|---|---|---|---|---|
| Resnet50 | 45.6 | 15.9 | 14.4 | 5.4 |
| Swin-T | 46.1 | 13.3 | 13.8 | 9.3 |
| Convnext | 38.8 | 12.4 | 22.5 | 4.5 |
| ViT-b-16 | 37.9 | 13.3 | 20.2 | 8.0 |
| Ours | 61.5 | 26.6 | **43.4** | **24.9** |
| Ours++ | **65.0** | **28.0** | 43.4 | 24.4 |

Table 4: Consistent Class and Pose estimation (CCP) results for different datasets. A prediction is considered correct if the angular error is lower than $\frac{\pi}{18}$ with a right class label prediction. Higher is better. RCNet exhibits a substantial performance improvement over all baselines.

each category (Equation 3) to estimate the object pose. The category achieving the minimal reconstruction loss is selected as the class prediction. Inference takes $\sim 0.8s$ on 8 RTX 2080Ti.

**Evaluation.** We evaluate the tasks classification, pose estimation, and Consistent Classification and Pose estimation (CCP). The 3D pose estimation involves predicting azimuth, elevation, and in-plane rotations of an object with respect to a camera. Following [41], the pose estimation error is calculated between the predicted rotation matrix $R_{pred}$ and the ground truth rotation matrix $R_{gt}$ as $\Delta(R_{pred}, R_{gt}) = \frac{\|\log m(R_{pred}^T R_{gt})\|_F}{\sqrt{2}}$. We use two common thresholds $\frac{\pi}{18}$ and $\frac{\pi}{6}$ (see Supplementary for $\frac{\pi}{6}$ results) to measure the prediction accuracy. We note that for a correct CCP, the model must estimate both the class label and 3D object pose correctly.

**Baselines.** We compare our RCNet to 4 baselines. We consider the pose estimation problem as a classification problem by using 42 intervals of $\sim 8.6°$ for each parameter (azimuth angle, elevation angle, and in-plane rotation). For each baseline, we replaced the original head with 2 smaller heads: one for object classification, and one for pose estimation. In order to make baselines more robust, we apply some data augmentation (i.e., scale, translation, rotation, and flipping) for each baseline during training. It is important to note that we do not apply any such augmentation for RCNet during training. We also compare to NeMo [36] for 3D pose estimation, which is also a render-and-compare approach. We trained it for each dataset using the publicly available code and obtained similar performances to the ones reported in [36] (i.e., NeMo-SingleCuboid). The baselines we compare to, have proven their robustness in OOD scenarios [23, 36].

### 4.2 Robust Object Classification

We first evaluate the performance of our model on independently and identically distributed data (IID). As the L0 (clean images) column of Table 1 shows, RCNet achieves a classification score of more than 99% on IID data, which is comparable to other classical approaches.

Furthermore, RCNet manages to robustly classify images in various out-of-distribution scenarios. From Table 1, we can see RCNet outperforms all other traditional baselines with around 5% accuracy on average for dif-

ferent levels of occlusions and with up to 30% accuracy boost for images under five different types of nuisances in OOD-CV.

We also evaluate robustness under different corruptions in Table 2. We observe that some baselines (e.g., ConvNext, ViT) perform better under some corruptions. We can easily explain that by the fact that they have been pre-trained on massive amounts of data, sometimes including these corruptions. As a result, these corruptions become IID and the comparison might not be completely fair. Despite this consideration, RCNet performs better on average compared to all baselines.

Based on these evidences, RCNet has made a **great improvement in OOD** scenarios while **maintaining cutting-edge accuracy for IID** data for classification. Finally, it is also worth noting that RCNet is **much more consistent** than all baselines (i.e., results' standard deviations are 15.8 and 9.7 for ViT and Ours, respectively). Independently of the nuisance's nature, RCNet tends to have consistent performances. We think that 3D-aware networks are less likely to learn shortcuts than traditional CNNs or Transformers.

### 4.3 Robust 3D Pose Estimation

As previously mentioned, due to its generative nature, RCNet is capable of performing 3D pose estimation. According to the results presented in Table 3, RCNet outperforms all feed-forward baselines significantly across all datasets. In addition, RCNet competes with NeMo, the current state-of-the-art method for robust 3D pose estimation. On average, RCNet performs better than NeMo across all datasets, except for occluded-Pascal3D+, where NeMo, designed specifically for occluded 3D pose estimation, holds a slight advantage. Despite not being explicitly designed for robust 3D pose estimation, we can conclude that RCNet achieves comparable performance to state-of-the-art architectures that are specifically tailored for this task.

### 4.4 Consistent Class and Pose estimation

RCNet can make consistent class label prediction and 3D pose estimation in a single passing. In Table 4, we can see that when considering accuracy for predictions with

| Metric | CCP@$ACC_{\frac{\pi}{18}}$ ↑ | | Computation cost ↓ (in % of the reference) | |
|---|---|---|---|---|
| Datatset | P3D+ | occluded-P3D+ | P3D+ | occluded-P3D+ |
| FC-Head | 45.6 | 15.9 | 3 | 3 |
| Ours | 61.5 | 26.6 | 100 | 100 |
| prev.+$S_1$ | 61.3 | 26.4 | 25 | 88 |
| prev.+$S_2$ | 63.8 | 24.7 | 8 | 29 |
| prev.+$S_3$ | **65.0** | **28.0** | 12 | 75 |

Table 5: Ablation of the performance and inference cost of our model components at Consistent Class and Pose Estimation (CCP) on P3D+ (L0) and occluded-P3D+ (L1-L3). FC-head refers to our trained backbone with a fully-connected-head. Ours refers to RCNet described in 3.2. We subsequently add the three stages of our proposal verification approach ($S_1$-$S_3$), where "prev." refers to previous line. Ours+$S_1$+$S_2$+$S_3$ is equivalent to Ours++ in other tables.

both right class label and low pose estimation error, RCNet **outperforms all baselines** by a large margin.

We believe, these enhanced capabilities come from the consistency between both tasks of object classification and 3D pose estimation. This illustrates that RCNet, unlike naive multi-head architectures, learns latent connections between object pose and object category. RCNet uses information from 3D poses to make classification more accurate, and it has turned out, both intuitively and experimentally, that such information is extremely helpful for enhanced robustness in the case of classification against OOD images.

In Figure 3, we provide qualitative results. Every image is OOD data with different nuisances. We can see that these scenarios are very likely to be encountered by classification models in the real world. From the correct mesh selected and the correct pose the mesh has aligned, and we can see how RCNet successfully predicts both the object category and object pose for these challenging images.

### 4.5 3D-aware head ablation

To assess the significance of our 3D-aware head, we conducted an ablation study by replacing it with a conventional fully connected head that predicts the object class and 3D pose. It is trained on the same data while freezing the backbone that has been trained as described in Sec-

tion 3.4. The results in Table 5 indicate that the conventional classifier is not as effective in utilizing the extracted features for Consistent Class and Pose estimation (CCP) as compared to our 3D-aware head model (referred to as "Ours"). The drop of 15.9% and 10.7% in unoccluded and occluded scenarios, respectively, demonstrates that the 3D-aware head is capable of interpreting the learned features in a more meaningful way. This finding once again suggests that FC models, and CNNs more broadly, tend to take some shortcuts during the learning process.

### 4.6 Speed-up by Proposal Verification

We seek to reduce the computational cost of our method while retaining the performance and robustness, as introduced in Section 3.2. Hence, we evaluate our proposal verification approach using two metrics: computational cost and accuracy. We analyze the effect of each component of our verification strategy separately for CCP with a threshold of $\frac{\pi}{18}$ (i.e., strictest task since a classification or pose estimation drop would induce a drop in CCP):

$S_1$: *Handle simple cases with FC-head.* We define the threshold $\tau_1 = 0.95$ experimentally to maximize the true positives while minimizing the false positives, i.e., the FC-head predicts the wrong category with high confidence. It allows us to reduce the processing requirements by 75% and 22% for the unoccluded and occluded subsets, respectively.

$S_2$: *Propose-and-verify with FC-head.* Using the FC-head output as initialization reduces computation by almost a factor of 4 for the occluded dataset. Furthermore, it is interesting to note that our proposed initialization scheme is very beneficial in unoccluded cases and leads to an improvement of more than 2% in accuracy while reducing computation by 10. We do not observe an accuracy gain for occluded images, which provides further evidence that **feed-forward models are less reliable in OOD scenarios**.

$S_3$: *Full render-and-compare when uncertain.* For 19% of the test images, the render-and-compare optimization does not converge to a good solution. By applying a threshold on the reconstruction loss $\tau_2 = 0.8$, we can recover 92% of these wrong predictions that the feed-forward models, at the cost of higher computation cost. These thresholds generalize well under different OOD scenarios and we study their sensitivity in the Supplementary.

In short, we observe a synergistic effect in many situations, leading to improved performance in terms of computation and even accuracy when combining standard (non-robust) fully-connected heads with our 3D-aware head.

## 5 Conclusion

In this work, we introduce RCNet, a 3D-aware neural network architecture that utilizes a render-and-compare approach for object classification. Our model represents objects as cuboid meshes with discriminatively trained neural textures and performs classification along with 3D pose estimation in a unified manner through inverse rendering. We integrate the 3D-aware head with a standard (non-robust) fully-connected head to considerably speed up the render-and-compare inference and reach a sweet spot that further enhances the performance while also greatly reducing the computational cost at inference time. Our experiments demonstrate that RCNet robustly classifies objects in a variety of OOD scenarios while also performing competitively at 3D pose estimation. We believe our work showcases the potential of integrating render-and-compare approaches and deep networks to achieve high performance with robustness in OOD scenarios. Furthermore, we strongly encourage future research to evaluate their methods on more challenging datasets to obtain better estimates of their performance in complex real-world scenarios.

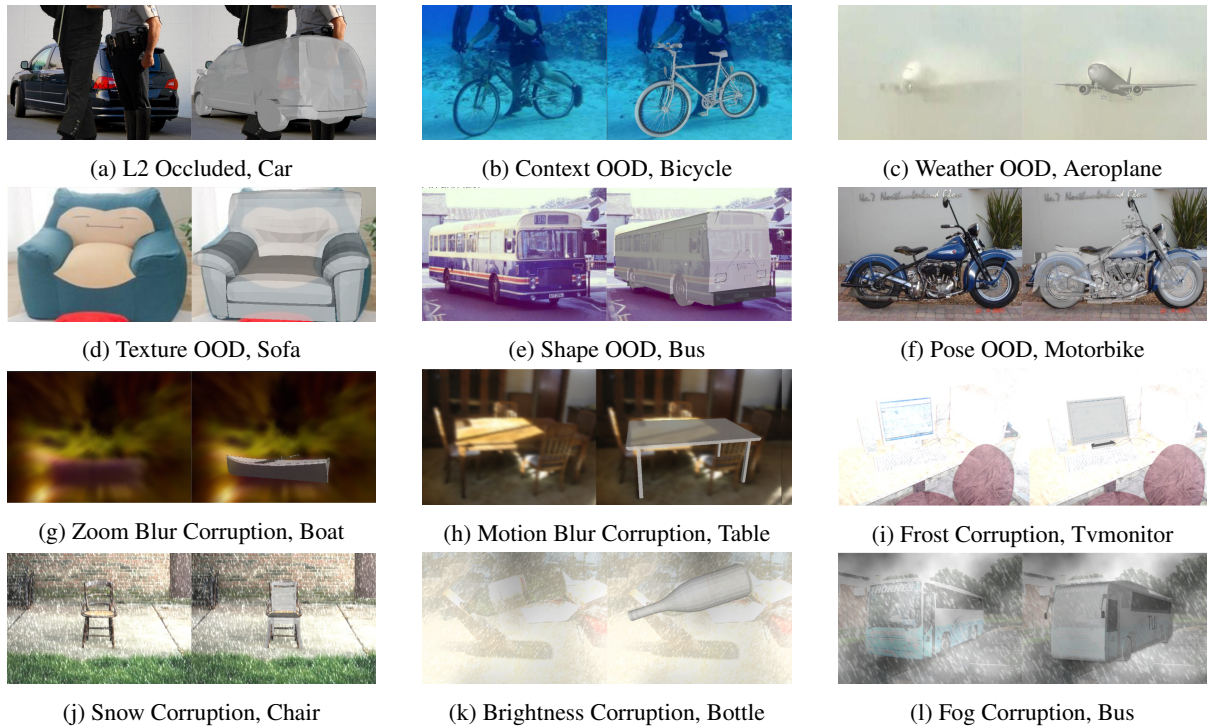

Figure 3: Qualitative results of RCNet on Occluded PASCAL3D+ and OOD-CV (a-f), and on Corrupted PASCAL3D+ (g-l). We illustrate the predicted 3D pose using a CAD model. Note that the CAD model is not used in our approach. All images were correctly classified by the proposed RCNet model but incorrectly classified by at least one baseline.



\end{document}